\documentclass[twocolumn,letterpaper]{IEEEtran}
\usepackage{multirow}
\usepackage{amsmath}
\usepackage{amssymb}
\usepackage{graphicx}
\usepackage{subfigure}
\usepackage{verbatim}
\usepackage{bookmark}
\usepackage[thinlines,thiklines]{easybmat}
\usepackage{latexsym}
\usepackage[dvipsnames]{xcolor}
\usepackage{cite}
\usepackage{stmaryrd}
\usepackage{multirow}
\usepackage{textcomp}
\usepackage[left=0.75in,top=0.75in,right=0.75in,bottom=0.75in]{geometry}
\usepackage{stfloats}
\usepackage{setspace}
\usepackage{algorithm}
\usepackage{algpseudocode}
\usepackage{algorithmicx}

\usepackage{booktabs}

\newsavebox{\twosubbox}

\newcommand{\bs}[1]{\ensuremath{{\boldsymbol{#1}}}}

\def\s{\mathop{\rm s}\nolimits}
\def\c{\mathop{\rm c}\nolimits}

\def\diag{\mathop{\rm diag}\nolimits}

\def\diag{\mathop{\rm diag}\nolimits}

\def\arctan2{\mathop{\rm arctan2}\nolimits}

\newcommand{\Notations}[1]{\item[]\noindent\ALG@special@indent \textbf{Notations:}\ #1}

\makeatletter
\newcommand{\multiline}[1]{%
	\begin{tabularx}{\dimexpr\linewidth-\ALG@thistlm}[t]{@{}X@{}}
		#1
	\end{tabularx}
}
\makeatother

\graphicspath{{./figures/}}
       
\title{Exoskeleton-Assisted Stance and Kneeling Balance and Work Task Evaluation in Construction~\thanks{The work was supported in part by US NSF under awards CMMI-2222880 and DGE-2021628. An earlier version of this paper was presented in part at the 2023 IEEE International Conference on Automation Science and Engineering, Auckland, New Zealand, August 26-30, 2023 [DOI: 10.1109/CASE56687.2023.10260384]. ({\em Corresponding author: Jingang Yi}).}}

\author{Gayatri Sreenivasan\thanks{G. Sreenivasan was with the Department of Electrical and Computer Engineering, Rutgers University, Piscataway, NJ 08854 USA. She is now with Elmore Family School of Electrical and Computer Engineering, Purdue University, West Lafayette, IN 47907 USA (email: gs912@scarletmail.rutgers.edu).}, Chunchu Zhu, and Jingang Yi\thanks{C. Zhu and J. Yi are with the Department of Mechanical and Aerospace Engineering, Rutgers University, Piscataway, NJ 08854 USA (email: chunchu.zhu@rutgers.edu; jgyi@rutgers.edu).}}

\begin{document}

\maketitle

\begin{abstract}
Construction workers experience serious safety and health risks in hazardous working environments. Quiet stance and kneeling are among the most common postures performed by construction workers during their daily work. This paper first analyzes lower-limb joint influence on neural balance control strategies using the frequency behavior of the intersection point of ground reaction forces. To evaluate the impact of elevation and wearable knee exoskeletons on postural balance and welding task performance, we design virtual- and mixed-reality (VR/MR) to simulate elevated environments and welding tasks. A linear quadratic regulator-controlled triple- and double-link inverted pendulum model is used for balance strategy quantification in stance and kneeling, respectively. Extensive multi-subject experiments are conducted to evaluate the usability of wearable exoskeletons in destabilizing construction environments. The quantified balance strategies capture the significance of the knee joint during balance control of quiet stance and kneeling gaits. Results show that center of pressure sway area is reduced up to 62\% in quiet stance and 39\% in kneeling gait for subjects tested in high-elevation worksites with knee exoskeleton assistance. The balance and multitask evaluation confirm and provide guidance on exoskeleton design to mitigate the fall risk in construction.
\end{abstract}

\begin{notetopractitioners}
Construction workers commonly perform tasks that require prolonged quiet stance or kneeling gaits on high elevations. Worker balance can be undermined by chronic knee injuries, musculoskeletal disorders, and destabilizing visual perturbations caused by occupational activities. Wearable knee exoskeletons have evolved as promising interventions to reduce knee joint stress across a variety of work gaits in construction. Emerging technologies such as virtual- and mixed-reality (VR/MR) provide an enabling tool to study underlying balance strategies to complete tasks in dynamic environments. The VR/MR- generated immersive elevated welding environment are leveraged to examine the effects of threatening visual stimuli, wearable exoskeletons, and construction tasks on worker balance and skill performance. Intersection point height frequency analysis is used to quantify the neural balance strategy during various testing scenarios. We particularly explore the often-neglected role of the knee joint to facilitate research on knee exoskeleton assisted balance in stance and kneeling gaits. The experimental results provide insight into the efficacy of knee exoskeletons in improving worker's stability and task performance in construction. The results highlight the need for a holistic approach to exoskeleton design to ensure that developed solutions can be safely and successfully integrated into the workplace environment.
\end{notetopractitioners}

\begin{keywords}
Postural balance, knee exoskeleton, virtual/mixed reality, automation in construction, biomechanics
\end{keywords}

\section{Introduction}
\label{intro}

Construction workers commonly perform tasks that require prolonged quiet stance or kneeling gaits in high elevations or cluttered environments. These awkward gaits in construction increase the risk of work-related musculoskeletal disorders (WMSDs), such as knee pain or knee osteoarthritis~\cite{palmer2012occupational}. In both stance and kneeling gaits, human workers are inherently unstable in an upright posture, requiring continuous engagement of muscles to regulate joint torques and maintain postural stability~\cite{WinterJNP1998,allen2016neuromechanical}. Although the neural balance mechanism at quiet stance has been extensively studied, the underlying control strategy remains elusive~\cite{boehm2019frequency}. Previous studies have investigated the ankle or hip strategies during quiet standing~\cite{winter2001ankle,ChenCASE2021}, and most of them used the single-link inverted pendulum model to analyze human balance strategies. The importance of the knee and hip joints at quiet stance was demonstrated in~\cite{YamamotoGP2015,gunther2009all}. The knee joint contribution to maintaining an upright stance is however often overlooked. Few research work was reported for postural control of kneeling gaits. In~\cite{MezzaraneEBR2008}, a comparative analysis of quiet kneeling and stance postures was conducted with a focus on the role of visual feedback during kneeling. A single inverted pendulum model was used in~\cite{ChenCASE2021} to study the influences of elevation on the postural balance of kneeling gait.

All of the works above primarily use the motion of the center of mass (COM) or center of pressure (COP) of the foot-ground contact to quantify the human balance performance. Metrics such as power spectral density (PSD) of the COP and acceleration were used to compare and discriminate between quiet stance and kneeling~\cite{MezzaraneEBR2008,ChenCASE2021}. The metrics solely derived from COP or COM are limited in capturing the balance dynamics and control as they are unable to account for both translational and rotational body accelerations simultaneously. To address these limitations, the study in~\cite{gruben2012mechanical,boehm2019frequency} introduced the intersection point (IP) of the ground reaction forces (GRF) during quiet stance. In~\cite{shiozawa2021frequency}, a double-link inverted pendulum (DIP) model was used to quantify the frequency-dependence of the IP height during quiet stance. The study focused on quantifying ankle and hip joint strategies and emphasized the importance of multiple joint-level contributions to overall balance control. 

Among the diverse environmental factors affecting balance control for construction workers, elevation is notably significant~\cite{BhattOERG2002,boffino2009fear,habibnezhad2019experiencing}. Acrophobia exacerbates this challenge, as it leads individuals to overestimate heights, intensifying fear and triggering physiological reactions that undermine balance~\cite{xing2019multicomponent}. Additionally, visual and environmental perturbations further compromise postural stability~\cite{luo2018effect}. To address these challenges, wearable devices like knee exoskeletons have emerged as pivotal fall risk interventions~\cite{AwolusiAIC2018,TrkovTASE2019,ChenTMECH2021,zhu2023knee}. However, few work has been reported on knee-based balance augmentation for stance and kneeling. Virtual- and mixed-reality (VR/MR) has been used to simulate high-elevation conditions and study neural balance control among construction workers~\cite{SimeonovHF2005,ChenCASE2021}. The work in~\cite{shi2019impact} showed safety reinforcement training on fall prevention for construction workers using VR scenes. The work in~\cite{habibnezhad2020neurophysiological} conducted neurophysiological assessments for welding task performance at virtual height to study cognitive effects on balance. A comprehensive review of VR/MR applications in construction safety can be found in~\cite{li2018critical}.

One main goal of this work is to study the influence of lower-limb joints, particularly the often-overlooked knee joint, on neural balance control in stance and kneeling gaits. We also evaluate the impact of elevation and wearable knee exoskeletons on postural balance and welding task performance in a realistic construction environment. Similar to previous studies (e.g.,~\cite{kuo1995optimal,Kooij1999,shiozawa2021frequency}), a linear quadratic regulator (LQR) is used as the human neural balance controller. Triple- and double-link inverted pendulum (TIP and DIP) models are taken to capture the joint-level effects for quiet stance and kneeling gaits, respectively. The models are used to quantify the observed neural balance control strategy from the frequency characteristics of the IP heights. The LQR allows for the explicit consideration of different parameters on balance performance, facilitating the interpretation of results in terms of physiological and biomechanical implications~\cite{shiozawa2021frequency}. Human subject experiments are conducted using incorporating VR/MR environments that simulate realistic welding worksites at varying elevations to evaluate the influence of elevation on postural balance and task performance. The subjects are outfitted with a wearable knee exoskeleton to test the usability and functionality of the intervention. The combination of experimental data and model analysis helps the extraction of physiologically plausible insights into the balance strategy chosen by subjects. Both objective metrics and human subjective feedback confirm that the exoskeletons have  promising potentials to improve both balance and task execution efficiency in construction.

The main contributions of this work are twofold. First, the study introduces new modeling and analysis for neural balance control that accounts for knee joint strategy with the IP height as the performance metric for both quiet stance and kneeling. To the authors' best knowledge, no previously reported work elucidates the involvement of knee joint strategy in stance or kneeling gaits. Second, this study provides a comprehensive evaluation of how elevation and wearable exoskeletons affect balance and task performance in construction. The quantitative findings offer valuable insights for the development of balance-enhanced wearable robotic interventions. This study integrates VR/MR to simulate realistic construction trades, which improves the accuracy of assessing exoskeleton-assisted balance and skilled task performance. Compared with the previous conference presentation~\cite{sreenivasan2023neural} that focused only on quiet stance, this work extends to kneeling gaits by introducing the TIP/DIP models to investigate the joint-level effects on postural balance control. The use of the LQR neural controller allows us to analyze the hip, knee, and ankle joints' contributions specifically to stance and kneeling balance. The analysis and comparison with experiments are new and contribute to a broader understanding of different work postures in construction.

The rest of the paper is organized as follows. We introduce the multi-link models and neural balance control in Section~\ref{model2}. The experimental setup and protocols are presented in Section~\ref{setup}. Data analysis and processing pipelines are explained in Section~\ref{data}. Section~\ref{results} presents the experimental  results, followed by discussions in Section~\ref{dis}. Finally, we give the concluding summary in Section~\ref{con}.

\begin{figure*}[ht]
	\hspace{-3mm}
	\subfigure[]{
		\label{TIPmodel}
		\includegraphics[width=3.58in]{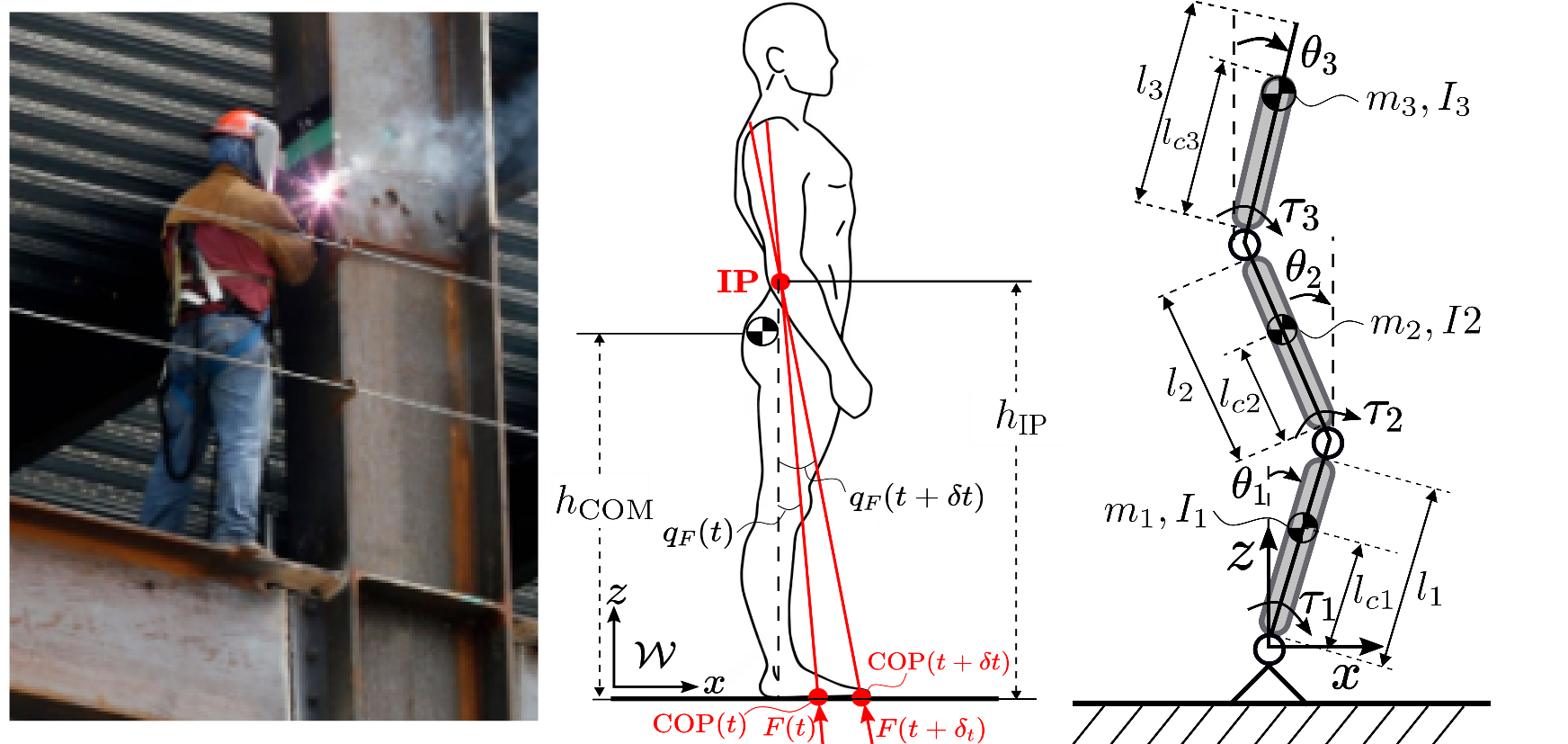}} 
	\hspace{-5mm}
	\subfigure[]{
		\label{IP_def}
		\includegraphics[width=3.58in]{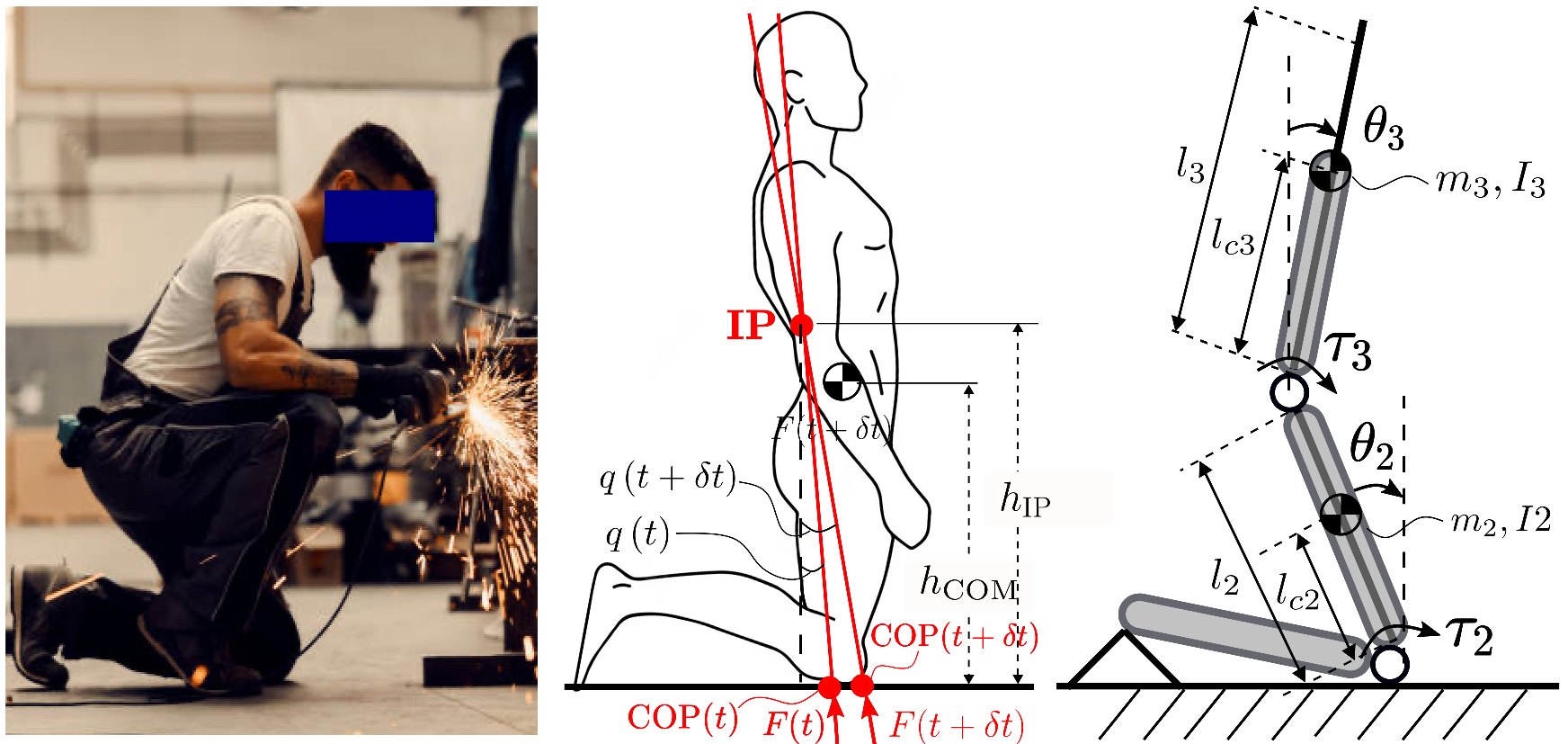}} 
\vspace{-1mm}
\caption{Stance and kneeling gaits in construction. (a) From left to right: Welder stands while conducting welding task; schematic of COM and IP in stance; schematic of the TIP model. (b) From left to right: welder kneels on both legs; schematic of the COM and IP in kneeling gait; schematic of the DIP model.}
\label{schematics}
\vspace{-2mm}
\end{figure*}

\section{Multi-Link Models and Balance Control}
\label{model2}

In this section, we first present the inverted pendulum models that describe quiet stance and kneeling, respectively. Next, the IP for human postural balance is discussed and we then present the LQR-based neural controller. 

\subsection{Biomechanical Models for Stance and Kneeling} 

Fig.~\ref{TIPmodel} illustrates the welder's stance posture and schematic consisting of three interconnected segments. The upper segment, which includes the head, arms, and trunk, is connected to the middle segment, representing the thigh, and the lower segment, representing the shank. Fig.~\ref{IP_def} shows the welder's working pose when kneeling on the ground with both legs alongside the schematic of the DIP model. In the case of quiet stance, the ankle joint is assumed to be the pin joint on the ground, whereas for kneeling, the knee joint is assumed to be the pin joint on the ground and feet are assumed to support no weights. Several assumptions are considered to model quiet stance and kneeling gaits. Small angle approximation is used by assuming that body movements remain small. We focus on planar motion in only the sagittal plane. The human body is modeled as a series of rigid segments (via DIP and TIP models). Additionally, we assume that human workers maintain a stationary posture around the unstable equilibrium with only the foot or knee in contact with the ground, in other words, we do not consider multi-contact postures such as when workers lean forward or use their hands for additional support.

The joint angles and joint torques at the ankle, knee, and hip are denoted as $\theta_i$ and $\tau_i$, $i=1,2,3$, respectively. The masses and mass moments of inertia about the centers of each link are denoted as $m_i$ and $I_i$, $i=1,2,3$, respectively. The distance between the mass center and the lower end of each link is denoted as $l_{c_i}$, $i=1,2,3$, respectively. Defining $\bs{\theta}_t=[\theta_1 \; \theta_2 \; \theta_3]^T$ and $\bs{\theta}_d=[\theta_2 \; \theta_3]^T$, the equations of motion for the TIP/DIP models are written as 
\begin{equation}
	\bs{M}_i(\bs{\theta}_i) \ddot{\bs{\theta}}_i + \bs{C}_i(\bs{\theta}_i,\dot{\bs{\theta}}_i) \dot{\bs{\theta}}_i +\bs{G}_i(\bs{\theta}_i)=\bs{\tau}_i, \; i=t,d, 
\label{equ0}
\end{equation}
where $\bs{M}_i(\bs{\theta}_i)$, $\bs{C}_i(\bs{\theta}_i,\dot{\bs{\theta}}_i)$, and $\bs{G}_i(\bs{\theta}_i)$ are the inertia, Coriolis, gravitational matrices, and subscripts ``t'' and ``d'' for the TIP and DIP models, respectively. The details of these matrices are given in~\cite{ScreenMS2023}. The human neural joint torques $\bs{\tau}_i$ are given as  
\begin{equation*}
\bs{\tau}_i= \bs{\tau}_{i}^s+\bs{w}_i, \; i=t,d,
\end{equation*}
where $\bs{\tau}^{s}_t = [\tau_1\; \tau_{2}\; \tau_{3}]^T$ and $\bs{\tau}^{s}_d = [\tau_2\; \tau_{3}]^T$, $\bs{w}_i \sim \mathcal{N}(\bs{0},\bs{\sigma}_i)$ are the joint torque perturbations that are modeled as independently white Gaussian noises with zero mean and variance $\bs{\sigma}_t = \diag(\sigma^2_1, \sigma^2_2, \sigma^2_3)$ and $\bs{\sigma}_d = \diag(\sigma^2_2, \sigma^2_3)$.

Introducing $\bs{x}_i = [\bs{\theta}_i^{T} \; \bs{\dot{\theta}}_i^{T}]^T$, $i=t,d$, from~\eqref{equ0}, we obtain
\begin{equation}
\bs{\dot{x}}_i = \begin{bmatrix} \dot{\bs{\theta}}_i \\
		\bs{M}_i^{-1}(\bs{\theta}_i)\left(\bs{\tau}_i-\bs{C}_i(\bs{\theta}_i,\dot{\bs{\theta}}_i)-\bs{G}_i(\bs{\theta}_i)\right) \end{bmatrix}.
\label{equ1}
\end{equation}
The position vectors of the COM are denoted as $\bs{r}^M_i = [x^{m}_i \, z^{m}_i]^T$, $i=t,d$, for stance and kneeling gaits, respectively. For stance posture, $x^{m}_t=-l_{c1}\s_{\theta_{1}} -l_{c2}\s_{\theta_{1}+\theta_{2}}-l_{c3}\s_{\theta_{1}+\theta_{2}+\theta_{3}}$ and $z^{m}_t=l_{c1}\c_{\theta_{1}}+l_{c2}\c_{\theta_{1}+\theta_{2}}+l_{c3}\c_{\theta_{1}+\theta_{2}+\theta_{3}}$, where notations $\s_{\theta_i}=\sin \theta_i$, $\c_{\theta_i}=\cos \theta_i$, $\s_{\theta_i+\theta_j}=\sin(\theta_i+\theta_j)$, and $\c_{\theta_i+\theta_j}=\cos(\theta_i+\theta_j)$ for angles $\theta_i$ and $\theta_j$, $i\neq j$, $i,j = 1,2,3$. For quiet stance, we define the Jacobian matrix of the COM to $\bs{x}_i$ as $\bs{J}^{M}_t=-[\bs{J}^{M_{1}}_t \; \bs J^{M_{2}}_t \;	\bs J^{M_{3}}_t]$, where $\bs{J}^{M_{1}}_t=\bs{r}_t^M$, $\bs J^{M_{3}}_t=[	l_{c3}\c_{\theta_{1}+\theta_{2}+\theta_{3}}\;l_{c3}\s_{\theta_{1}+\theta_{2}+\theta_{3}}]^T$, and 
\begin{equation*}
\bs{J}^{M_{2}}_t =\begin{bmatrix}
		l_{c2}\c_{\theta_{1}+\theta_{2}}+l_{c3}\c_{\theta_{1}+\theta_{2}+\theta_{3}} \\
		l_{c2}\s_{\theta_{1}+\theta_{2}}+l_{c3}\s_{\theta_{1}+\theta_{2}+\theta_{3}}
	\end{bmatrix}.
\end{equation*}
For kneeling gait, $x^{m}_d=-l_{c2}\s_{\theta_{2}}-l_{c3}\s_{\theta_{2}+\theta_{3}}$ and $z^{m}_d=l_{c2}\c_{\theta_{2}}+l_{c3}\c_{\theta_{2}+\theta_{3}}$. The Jacobian matrix of the COM is $\bs{J}^{M}_d= -[\bs{J}^{M_{2}}_d \; \bs{J}^{M_{3}}_d]$, where $\bs{J}^{M_{3}}_d=[ l_{c3}\c_{\theta_{2}+\theta_{3}}\;l_{c3}\s_{\theta_{2}+\theta_{3}}]^T$ and 
\begin{equation*}
\bs{J}^{M_{2}}_d=\begin{bmatrix} l_{c2}\c_{\theta_{2}}+l_{c3}\c_{+\theta_{2}+\theta_{3}} \\
		l_{c2}\s_{\theta_{2}}+l_{c3}\s_{\theta_{2}+\theta_{3}}
	\end{bmatrix}.
\end{equation*}
In the above model for quiet stance, $l_{ci}=\frac{\frac{1}{2}m_{i}l_{i}+\sum_{j=i+1}^{3}m_{i}l_{j}}{m^t_t}$, $i=1,2,3$, and $m^t_t = m_{1} + m_{2} + m_{3}$. For kneeling gait, $l_{ci}=\frac{\frac{1}{2}m_{i}l_{i}+\sum_{j=i+1}^{3}m_{i}l_{j}}{m^t_d}$, $i=2,3$, $m^t_d = m_{2} + m_{3}$ . The acceleration of the COM is obtained as $\bs{\ddot{r}}^{M}_i =[\bs{\dot{J}}^{M}_i \; \bs{J}^{M}_i]\dot{\bs{x}}_i=\mathcal{J}_i^M \dot{\bs{x}}_i$, where $\mathcal{J}_i^M =[\bs{\dot{J}}^{M}_i \; \bs{J}^{M}_i]$, $i=t,d$. The GRFs in the horizontal and vertical directions are computed as
\begin{equation}
F_{x}^i = m^t_i \ddot{x}^{m}_i, \quad F_{z}^i = m^{t}_i(\ddot{z}^{m}_i + g), \,\, i=t,d,
\label{eq11}
\end{equation}
where $g = 9.8$~m/s$^2$ is the gravitational constant, 

\subsection{Intersection Point}

The IP is the point in space where the net GRF vectors at adjacent time steps intersect. Fig.~\ref{schematics} illustrates the IP concept for quiet stance and kneeling. The IP height, denoted by $h_\text{IP}$, is calculated using the COP and GRF data obtained from force plate measurements. The angle between the GRF vector and the vertical direction is denoted as $q(t)$ at time $t$; see Fig.~\ref{schematics}. Assuming subtle movements of the body and using a small angle approximation, we obtain
\begin{equation}
	q(t) \approx  \tan (q(t)) = \left|\frac{F_x(t)}{F_z(t)}\right| ,
\label{eq22}
\end{equation}
where $\bs{F}(t)=[F_x(t) \; F_z(t)]^T$ is the GRF vectvor at $t$. It is clear that the relation between unfiltered $q(t)$ and COP is not linear. To calculate $h_\text{IP}$, we consider the COP position along the $x$-direction, denoted by $x_\text{CP}(t)$, at time moments of $t$ and $t+\delta t$. From Fig.~\ref{schematics}, we obtain
\begin{equation}
q(t)=\frac{x_\text{CP}(t)}{h_\text{IP}(t)}, \; q(t+\delta t)=\frac{x_\text{CP}(t+\delta t)}{h_\text{IP}(t)}.
\label{eq0}
\end{equation}
We solve for $h_\text{IP}(t)$ from~\eqref{eq0} and obtain
\begin{equation}
	h_\text{IP}(t) = \frac{x_\text{CP}(t+\delta t) - x_\text{CP}(t)}{q(t) - q(t+\delta t)}.
\end{equation}

Unlike traditional methods such as COP or COM analysis, the IP model accounts for both translational and rotational accelerations of the body, providing a comprehensive perspective on balance control strategies. By incorporating both aspects of body motion, IP analysis offers a holistic understanding of how individuals maintain postural stability. Specifically, IP height across different frequencies serves as a crucial metric for quantifying neuromuscular control efforts~\cite{boehm2019frequency}. The frequency analysis is essential because balance control involves movements at different speeds. Low-frequency components reflect active, slower, and large-amplitude adjustments like swaying to maintain posture, while high-frequency elements capture passive, rapid, and small-amplitude corrections in response to quick perturbations~\cite{hufschmidt1980some}. Analyzing these different frequency components helps understand how the body's control mechanisms shift. 

IP height is directly related to this frequency analysis because it changes systematically across different frequency bands~\cite{boehm2019frequency}. At low frequencies, the IP height is typically above the COM, indicating that the control strategy emphasizes active rotational stability through the hip and ankle joints. This corresponds to slower, larger adjustments in posture, where the body consciously engages in active neuromuscular control, such as muscle co-contractions and sensory feedback, to counterbalance significant deviations from an upright stance. As the frequency increases, the IP height moves below the COM. This shift reflects a change in control strategy toward passive, rapid, fine-tuned adjustments, primarily through the lower body. At these higher frequencies, the body relies less on active control and more on the passive mechanical properties of the muscles and joints, such as joint stiffness and elasticity, to manage small, quick perturbations and maintain stability.

\subsection{Neural Balance Control}
\label{NBC}

We linearize~\eqref{equ1} around the upright equilibrium, that is, $\bs{x}^e_i=\bs 0$ under input $\bs{\tau}^{e}_i=\bs 0$, and obtain the linearized system
\begin{equation}
\dot{\bar{\bs x}}_i = \bs A^{l}_i\bar{\bs x}_i+\bs B^{l}_i\bar{\bs \tau}^s_i+\bs B^{l}_i\bs w_i, \, i=t,d,
\label{equ2}
\end{equation} 
where $\bar{\bs x}_i=\bs{x}_i-\bs x^{e}_i$, $\bar{\bs \tau}^s_i=\bs \tau^s_i-\bs \tau^{e}_i$, and $\bs A^{l}_i$ and $\bs B^{l}_i$ are the state and input matrices, respectively. The cost function is given as 
\begin{equation}
	\label{lqr}
	J_i = \int_{0}^{\infty} \left[\bar{\bs x}_i^{T} \bs Q_i \bar{\bs x}_i+ (\bar{\bs{\tau}}_i^s)^T \bs R_i \bar{\bs \tau}_i^s\right]dt,
\end{equation}
where positive symmetric definite matrices $\bs Q_i$ and $\bs R_i$ penalize the state and control input, respectively. The solution to~\eqref{lqr} is $\bs \tau^{s}_i=-\bs K^{L}_i\bs x_i$, where $\bs K^{L}_i$ is the gain matrix that is solved by the Riccati equation. Matrix $\bs R_i$, $i=t,d$, is selected and designed as
\begin{equation}
\bs R_i =\alpha_i \bs{\beta}_i, 
\label{equ4}
\end{equation}
where $\alpha_i$ is the parameter related to overall control effort, and $\bs{\beta}_i$ describes the relative magnitude of joint efforts. Here $\bs{\beta}_t=\diag(\beta_{1},\beta_{2},\beta_{3})$  for the stance controller, and $\bs{\beta}_d=\diag(\beta_{2},\beta_{3})$ for the kneeling controller. $\beta_{1}$, $\beta_{2}$, and $\beta_{3}$ represent effort at the ankle, knee, and hip joints, respectively. The LQR neural control guarantees stability if $(\bs A^{l}_i, \bs B^{l}_i)$ is controllable, and $\bs Q_i$ is chosen as an identity matrix to equally penalize each state's deviation.

Following observations from~\cite{shiozawa2021frequency}, the physiological effects of balance control strategies at the joint level can be inferred from the LQR-based neural controller controller parameters $\alpha_i$, $\bs{\beta}_i$, $\bs{\sigma}_i$. Specifically, $\alpha_i$ characterizes the control effort employed to attain stability, $\bs{\beta}_i$ represents the contribution of ankle, knee, and hip joints to balance control, and $\bs{\sigma}_i$ captures the neuromuscular impedance and noise at each joint. 

\begin{figure*}[ht!]
	\centering
		\includegraphics[width=6.8in]{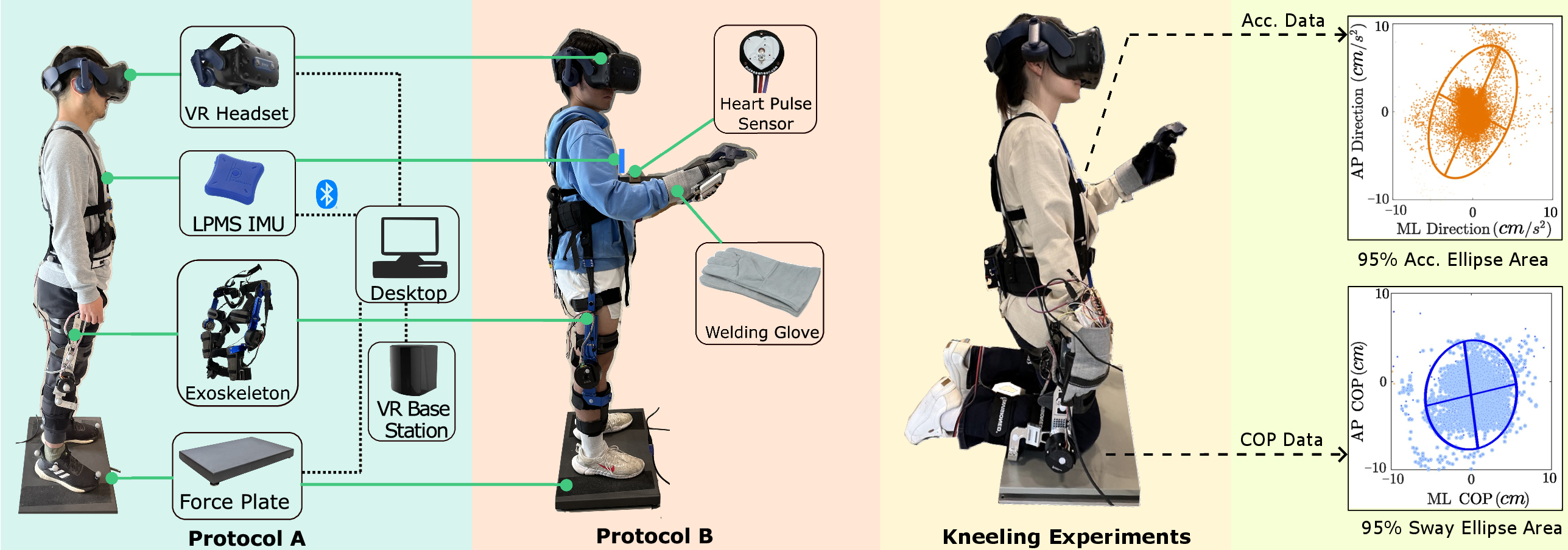}
	\caption{The VR/MR-enhanced human subject balance experiment setup with wearable knee exoskeletons for Protocols A (left) and B (mid), and an explanation for the kneeling experiments and the $95$\% acceleration and sway ellipse area (right). }
	\label{fig:exp_setup}
	\vspace{-2mm}
\end{figure*}

The output of~\eqref{equ2} is considered as $y_{1i}=-F_x^i$, $i=t,d$, and $y_{2t} = \tau_1$ (ankle torque for stance) or $y_{2d} = \tau_2$ (knee torque for kneeling) that are directly measured from force plate. The negative sign in $y_{1i}$ is due to the convention that a positive horizontal GRF is directed towards the positive $x$-axis; see Fig.~\ref{TIPmodel}. From~\eqref{eq11} and using the Jacobian matrix, we obtain 
\begin{equation*}
	y_{1i} =-m^t_i\ddot{x}_i^m=-m^t_i[1\; 0] \mathcal{J}^{M}_i\dot{\bs{x}}_i=\bs{C}_i \bs{x}_i+\bs{D}_i^1 \bs{\tau}_i^s,
\end{equation*}
where $\bs C_{i} = -m_{t}[1 \; 0]\mathcal{J}^{M}_i\bs A^{l}_i$ and $\bs{D}_{1i}=-m_{t}[1 \; 0]\mathcal{J}^{M}_i\bs B^{l}_i$. It is clear that $y_{2i}=\bs{D}_{2i}\bs{\tau}_i^s$, where $\bs{D}_{2i}=[1 \; \bs{0}_i]$, $i=t,d$, and $\bs{0}_t=[0 \; 0]$ and $\bs{0}_d={0}$. Therefore, output equation is 
\begin{equation*}
	\bs y_i = \begin{bmatrix} y_{1i} \\ y_{2i} \end{bmatrix}= \begin{bmatrix}	\bs C_i \\\bs{0} \end{bmatrix}\bs{x}_i+\begin{bmatrix} \bs{D}_{1i} \\ \bs D_{2i}	\end{bmatrix}\bs \tau_i^s, \, i=t,d.
\end{equation*}

\begin{figure*}[ht!]
	\centering
	\subfigure[]{
		\includegraphics[width=3.38in]{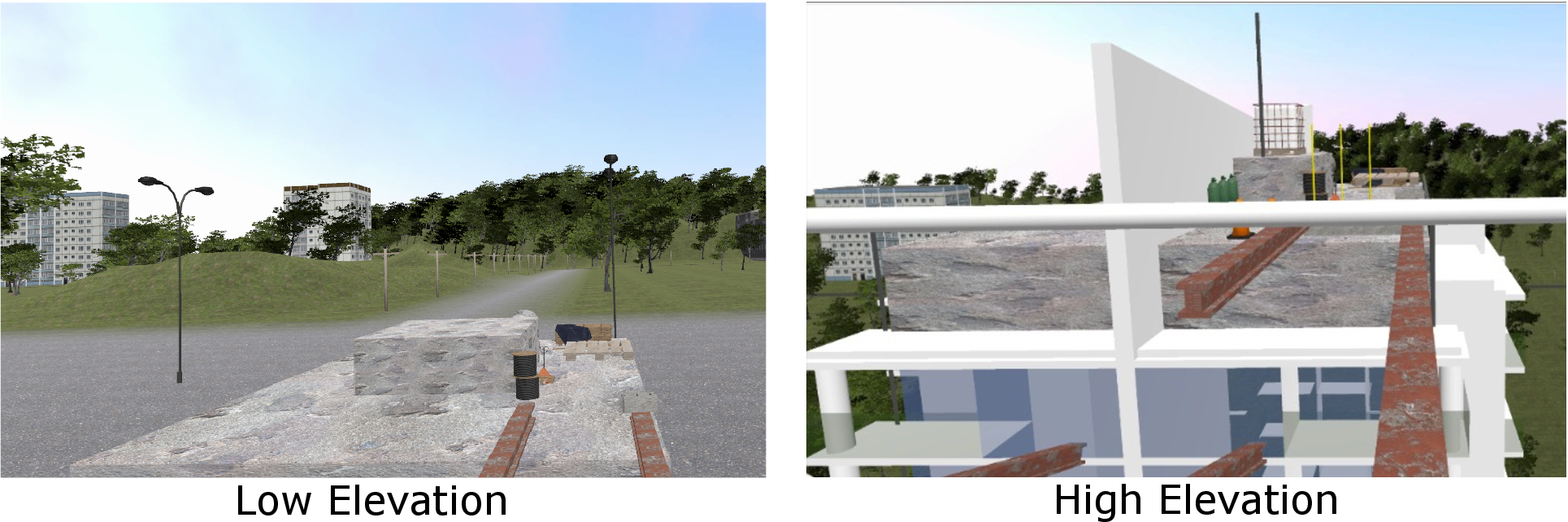}
		\label{fig:VR}
	} 
	\hspace{-1mm}
	\subfigure[]{
		\includegraphics[width=3.35in]{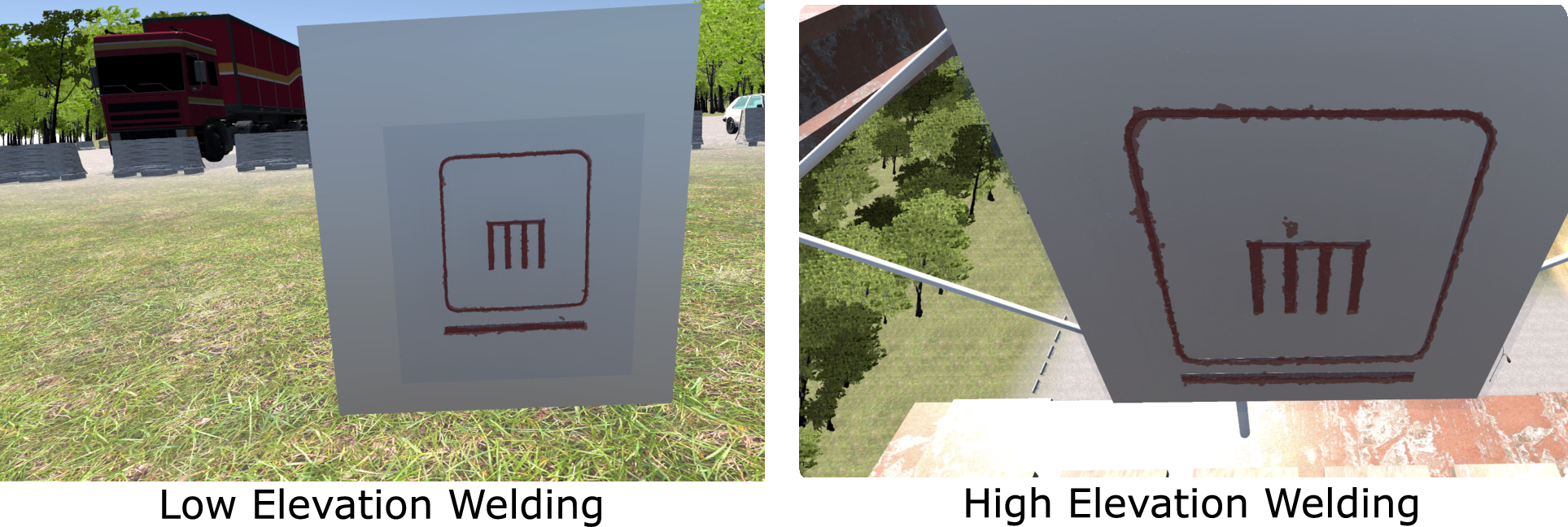}
		\label{fig:WeldingScene}
	} 	
	\vspace{-3mm}
	\caption{(a) VR scene in Unity engine at LE (left) and HE (right) views. (b) VR view in the welding tasks at LE ground view (left) and HE view (right).}
	\label{VRview}
	\vspace{-3mm}
\end{figure*}

\section{Experiments}
\label{setup}

\subsection{Experiment Setup and Protocol}

Fig.~\ref{fig:exp_setup} shows the experimental setup for quiet stance and kneeling gaits in construction. A wearable inertial measurement unit (IMU) (from LP-RESEARCH Inc.) was attached to the subject's chest to measure postural sway accelerations. A force plate (from Bertec Corporation) was used to collect GRF and COP data. A heart rate sensor (from World Famous Electronics LLC.) was used to capture the subject's heart rate during experiments. The human subjects were also equipped with bilateral knee exoskeletons to provide assistive torques and distribute musculoskeletal load and stress from the knee joint to the thigh and shank. The exoskeleton’s weight is supported by a waist belt, which is connected to the thigh support frame through elastic straps. The belt also contains an integrated compartment for the microcontroller and battery. The elastic straps secure the knee actuator, ensuring proper alignment. The thigh support frame includes a height-adjustable aluminum linkage on the lateral side of the leg, along with cuffs positioned on the back of the upper thigh and front of the lower thigh. The knee joint actuation system possessed high-torque, high-backdrivability, and high-bandwidth features~\cite{YuTMech2020}. 
	
The portable knee exoskeleton features a compact and lightweight design (unilateral unit weighed only $1.7$~kg, including electronics and battery); see Fig.~\ref{fig:exp_setup}. Each leg has a quasi-direct drive actuator ($35$~Nm peak torque) to provide a large range of motion ($0$-$160$~deg flexion), and high rotation speed ($16.2$~rad/s). The exoskeleton torque control was implemented on a low-level microcontroller (from PJRC Teensy) via CAN bus. All sensor data were synchronized and collected at 100~Hz using custom Python scripts running on a portable high-performance microcomputer (model NUC7i7DNK, Intel Corp.) The system ensures precise time-stamping of sensor data, enabling accurate synchronization during post-processing and analysis.

A VR system (HTC Vive Pro) was employed to immerse the subjects in diverse visual environments during test trials. A virtual construction site was built with the Unity game engine to simulate the typical low-elevation (LE) and high-elevation (HE) construction work environments. Fig.~\ref{fig:VR} shows the VR viewpoint for the quiet stance and Fig.~\ref{fig:WeldingScene} shows the VR scene for welding tasks. The LE scenario depicted the subject standing on a small platform slightly elevated from the ground. The HE scenario on the other hand simulated the subject standing atop a $30$-story building under ongoing construction. 

Eleven able-bodied human subjects ($4$ females, $7$ males, weight: $67\pm10$~kg, height: $170\pm8$~cm, age: $25\pm 2.5$~years) were recruited in the study. All participants provided informed consent and self-reported being in good health. The experimental protocol was approved by the Institutional Review Board (IRB) at Rutgers University. 
This study's experimental protocol encompasses two gait situations: quiet stance and kneeling. Each gait experiment is subdivided into two protocols, labeled Protocols A and B. The subjects were labeled from S1 to S11. The stance experiments involved the participation of $2$ female and $5$ male subjects, while the kneeling experiments were conducted with a group of $2$ female and $4$ male subjects. Subjects S6 and S7 participated in both stance and kneeling experiments.

\subsubsection{Protocol A - Posture balance evaluation}

Subjects were asked to perform a quiet stance and kneeling task while maintaining balance in a simulated environment for $60$~s. Subjects were given motion training to mimic construction worker gaits based on consultations with construction professionals. No additional contact points such as hands or other body parts were allowed during the experiment tasks. The visual stimuli varied between trials by altering the LE and HE. This setup facilitated two distinct trials, each designed to assess the impact of elevation perception on balance. The wearable knee exoskeleton was introduced in a subsequent variation. 

The results in~\cite{SREENIVASAN2023163} have shown that the knee joint's deviation from a natural stance is small and a stiffness-based controller effectively reduces muscle fatigue and enhances balance. Consequently, a stiffness knee torque controller $\tau_e = -k_r \theta_2$ was used for the quiet stance, where $\tau_e$ is the exoskeleton output torque, $k_r$ represents the reference joint stiffness, and $\theta_2$ denotes the knee joint's deviation from a natural stance. In contrast, during kneeling, the knee joint acts as a pin joint, bearing a significant portion of the body's weight. This increased loading on the knee joint leads to large variations in joint angles and requires great effort to maintain balance. Inspired by the work in~\cite{zhu2022wearable}, the inertia and Coriolis terms in~\eqref{equ0} are neglected and a proportional-derivative (PD) knee assistive torque controller with gravity compensation is considered  
\begin{equation*}
\tau_e=-k_p \theta_2 - k_d \dot{\theta}_2 -\gamma g\left[\frac{m_3}{2}\left(l_2 s_{\theta_2}-l_{c3} s_{\theta_3}\right) + m_2 l_{c2} s_{\theta_2}\right],
\end{equation*}
where $k_p$ and $k_d$ are the proportional and derivative gains, respectively, and $\gamma \in(0,1)$ is the assistance weight factor. 

Subjects were required to complete four trials in Protocol A. The trials, termed test of interest (TOI), incorporated both elevation scenarios with and without the exoskeleton support: (i) TOI~1: LE without the exoskeleton; (ii) TOI~2: HE without the exoskeleton; (iii) TOI~3: LE with the exoskeleton; and (iv) TOI~4: HE with the exoskeleton. A five-min rest period was given between each trial to minimize fatigue and ensure consistent performance.

\subsubsection{Protocol B - Welding task assessment}

A welding trade was considered as welders often have to hold awkward gaits while conducting tasks with precision requirements. The dual task of maintaining upright balance and performing welding accurately is challenging. Protocol B introduced a welding task in the VR scene (see Fig.~\ref{fig:WeldingScene}) and generated a dynamic and task-focused assessment. As shown in Fig.~\ref{fig:exp_setup}, participants were equipped with a VR controller in their dominant hand to emulate a welding gun, and were  asked to weld a workpiece as accurately as possible within the VR environment. To enhance realism, welding gloves were provided and no time constraints were imposed. The welding task was visually represented by ink emitted from the weld gun tip upon contact with the workpiece, resulting in a visual weld line. The completed works were documented as screenshots for subsequent analysis.

Similar to Protocol A, subjects underwent quiet stance and kneeling with four TOIs in Protocol B: (i) TOI~5: Welding task in an LE without the exoskeleton; (ii) TOI~6: Welding task in an HE without the exoskeleton; (iii) TOI~7: Welding task in an LE with the exoskeleton; and (iv) TOI~8: Welding task in an HE with the exoskeleton. Participants were given test trials to familiarize themselves with the welding setup and operation. They were instructed to perform balance and welding tasks exclusively in quiet stance and kneeling gaits, without adopting any other postures. Participants were allowed a $10$-min rest period between each trial to prevent task-induced fatigue and ensure accurate task execution. In both Protocols A and B, the orders of the trials were randomized for evaluation purposes.

In this study, participants wore the exoskeleton for approximately $60$~min across various tasks and experimental conditions. Although individual trials were relatively short, with each lasting around $60$~s, the cumulative exposure time allowed for the assessment of the device's comfort and usability beyond initial wear effects. Participants were provided with rest breaks between trials to minimize fatigue. To evaluate safety and comfort, both quantitative measures (e.g., postural sway, task performance, and physiological data such as heart rate) and qualitative feedback were gathered through post-experiment questionnaires, where subjects provided input on their comfort levels and overall experience with the exoskeleton.

\subsection{Data Collection and Processing Methods}

During the experiments, the $x$- and $y$-axis were respectively defined as anterior-posterior (A-P) and medial-lateral (M-L) directions. Fig.~\ref{fig:exp_setup} illustrates two key measurements used in analysis to quantify movement variability and postural stability: the $ 95\%$ acceleration ellipse area $A_i^{\text{Acc}}$ and the $95\%$ sway ellipse area $A_i^{\text{Sway}}$, $i=t,d$. $A_i^{\text{Acc}}$ is calculated as the elliptical area that contains 95\% of the acceleration variations (from IMU) in the A-P ($\ddot{x}^m_i$) and M-L ($\ddot{y}^m_i$) planes. It quantifies the variability in body accelerations and thus reflects the dynamic control and execution of movement. In contrast, $A_i^{\text{Sway}}$ is obtained as the elliptical area that contains $95\%$ COP data (from force plate) in the A-P ($x_{\text{CP}}$) and M-L ($y_{\text{CP}}$) directions. It represents the spatial variability of the COP movements, offering insights into the stability of postural equilibrium and the subject's ability to maintain or regain balance.

\begin{figure}[h!]
	\centering
	\includegraphics[width = 3.3in]{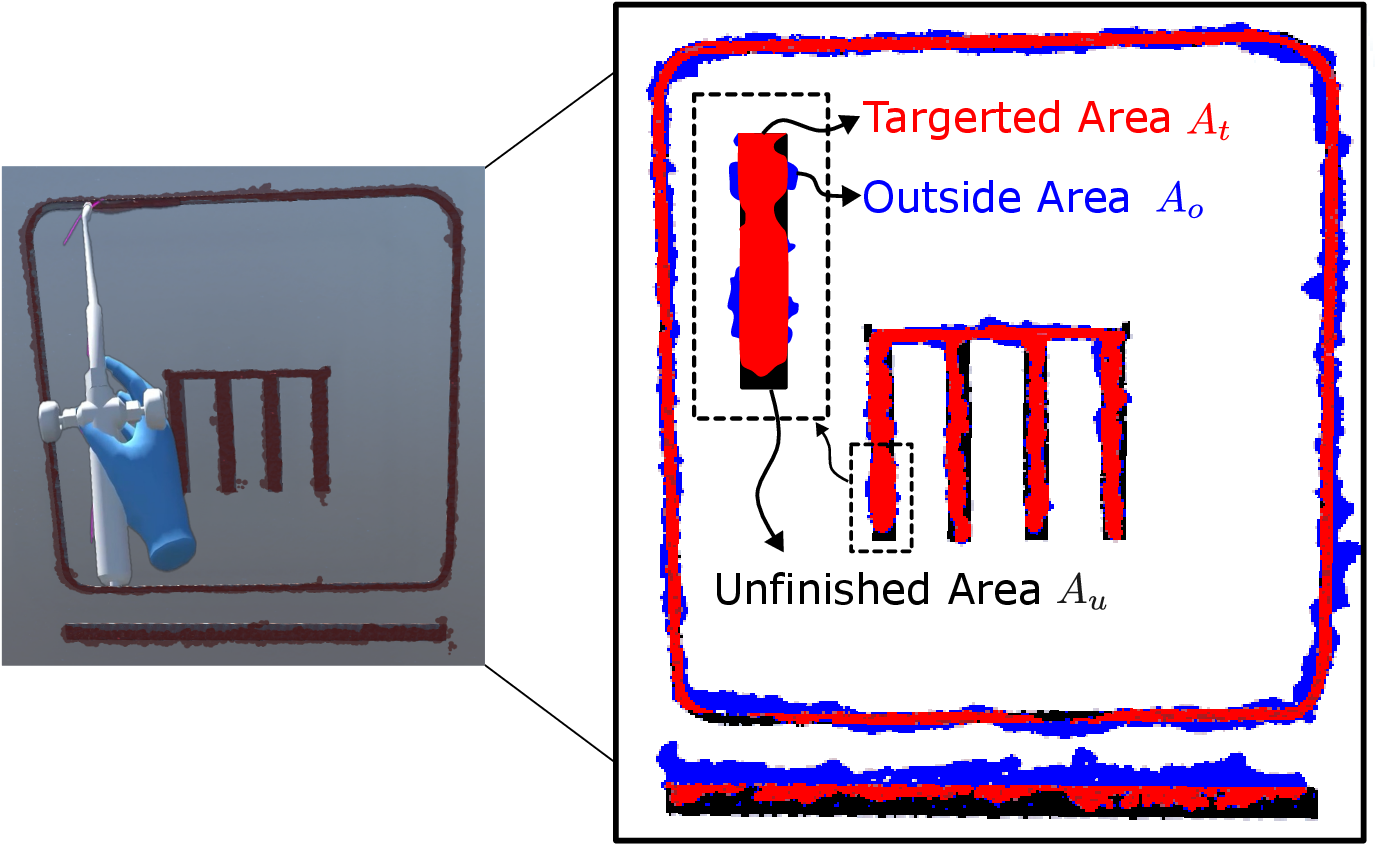}
	\vspace{-0mm}
	\caption{Image processing method explanation for welding performance assessment on a virtual welding piece.}
	\label{fig:imageProcessing}
\end{figure}

To evaluate welding task performance, digital image processing was used to analyze the images of the workpiece. Fig.~\ref{fig:imageProcessing} illustrates the extraction of correctly targeted welded area ($A_t$), outside area ($A_o$), unfinished area ($A_u$), and the total workpiece area ($A_{wp}$) to determine the accuracy of the welding efforts and to evaluate any unfinished segments. Three metrics were introduced to quantify the welding performance: accuracy, precision, and completion rate. Accuracy is defined as the ratio of the correctly welded area to the total workpiece area, i.e., $\frac{A_t}{A_{wp}}$. Precision measures the amount of welding that falls outside the targeted area, that is, $\frac{A_o}{A_{w}}$. Completion rate is assessed by the percentage of the workpiece that has been welded, that is, $\frac{A_{wp} - A_u}{A_{wp}}\times 100$\%. 

Subjective perceptions were also collected by pre- and post-experiment questionnaires (provided in the supplemental materials). Pre-experimentally, subjects' inherent fear responses were quantified using the James Geer Fear (JGF) questionnaire~\cite{geer1965development}, and the Cohen Acrophobia (CA) questionnaire~\cite{cohen1977comparison}, aimed at understanding their reactions to heights. Both questionnaires used a five-point Likert scale~\cite{joshi2015likert}. A subsequent post-experiment questionnaire sought to elucidate subjects' experiences within the VR scene, their perceptions of the experimental procedures, and their comfort and experience with the exoskeleton. This comprehensive questionnaire approach aimed to refine and enhance the experimental design through iterative feedback, while also gathering specific insights on the usability of the wearable device. This dual approach, melding subjective feedback with objective data, provides a holistic understanding of the impact of experimental conditions on both the psychological and physiological dimensions of participant response, as well as the practicality and user acceptance of the exoskeleton technology in the experimental context.

\section{Data Analysis and Simulation}
\label{data}

\begin{figure}[h!]
	\vspace{2mm}
	\centering
	\includegraphics[width=3in]{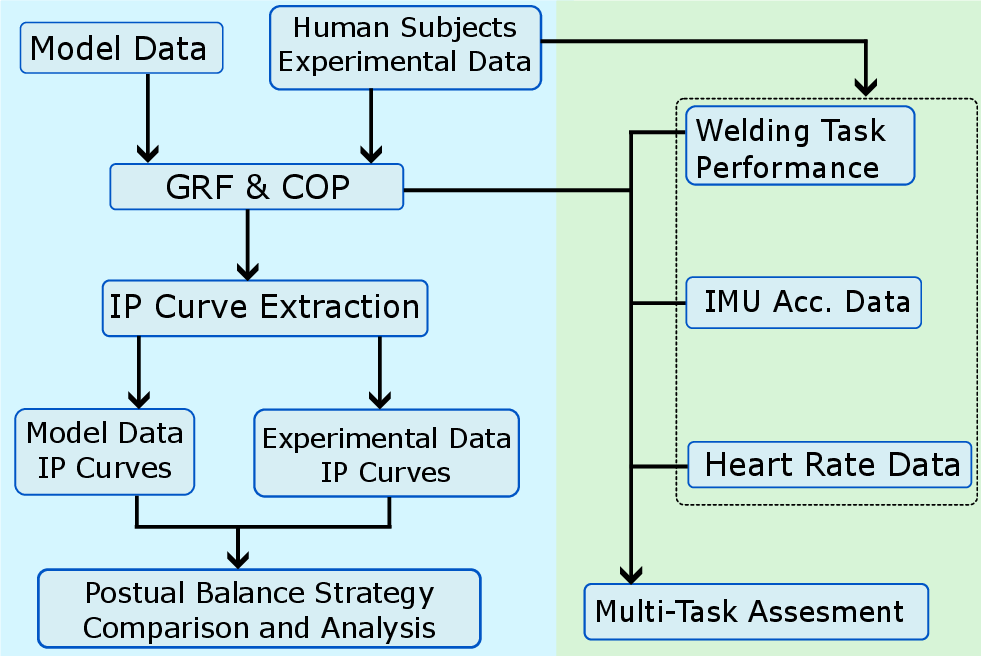} 
	\vspace{-2mm}
	\caption{Pipeline of intersection point height analysis and multi-task assessments. }
	\label{IPflow}
	\vspace{-1mm}
\end{figure}

Fig.~\ref{IPflow} shows the overall procedure for IP-based data analysis and multi-task assessment. The GRF and COP data were first derived from the experimental and model simulation data. To obtain the frequency curve for the $h_\text{IP}$, the COP and GRF data in the A-P direction were first processed using a Hann window. The windowed signals were then bandpass filtered using a zero-lag, 2nd-order Butterworth filter into $38$ non-overlapping frequency bands of $0.2$~Hz width, centered from $0.5$ to $7.9$~Hz. Parsing the signals into these bands reveals an approximately linear relationship between the COP and GRF data within each frequency band. The reciprocal of the slope of this linear trace represents the $h_\text{IP}$ for the corresponding frequency, as shown in~\eqref{eq22} and~\eqref{eq0}. 

\begin{figure}[h!]
	\centering
	\includegraphics[width=3.3in]{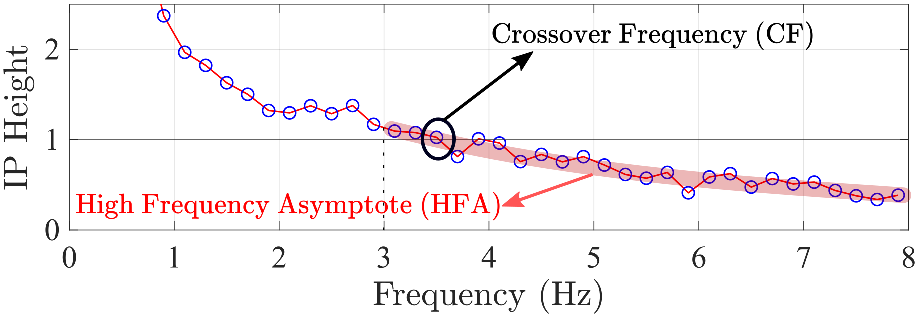} 
	\vspace{-1mm}
	\caption{Illustration of the crossover frequency $\omega^{\text{CF}}$ and the high-frequency asymptote (HFA) with slope $k^{\text{HFA}}$ of the IP height frequency curve.}
	\label{IPcurve}
	\vspace{-0mm}
\end{figure}

Fig.~\ref{IPcurve} shows an example of the IP frequency curve normalized by $h_{\text{COM}}=z^m_i$. We focus on the crossover frequency (CF) of the IP height frequency curve with the COM height (i.e., ratio $h_{\text{IP}} / h_{\text{COM}}$ at one), denoted by $\omega^{\text{CF}}$, and the slope of the high-frequency asymptote (HFA), denoted by $k^{\text{HFA}}$. To determine and calculate $k^{\text{HFA}}$ from the IP curves, an exponential fit was applied. For kneeling, the IP generally lies above the COM and it is challenging to define the CF and HFA in the same way as in stance. To maintain consistency in the analysis of balance control strategies, we introduce the concept of virtual CF and HFA by normalizing the IP height as a fraction of the upright stance COM height; see Fig.~\ref{IP_def}. 
	
\begin{figure*}[th!]
	\hspace{-4mm}
	\subfigure[]{
		\label{IP_CASE:a}
		\includegraphics[width=1.8in]{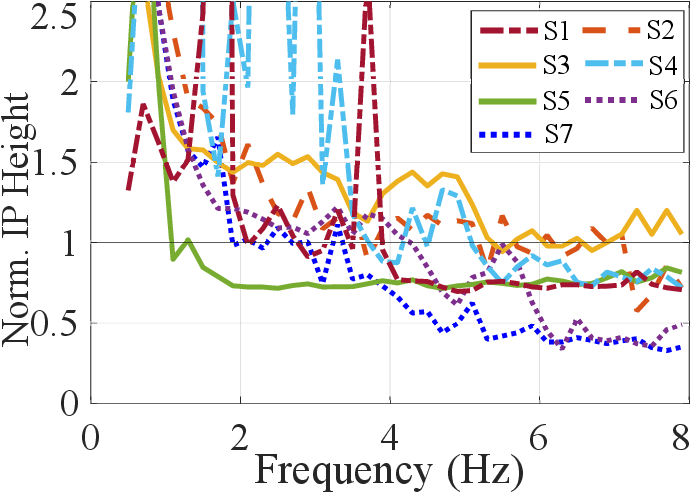}}
	\hspace{-3mm}
	\subfigure[]{
		\label{IP_CASE:b}
		\includegraphics[width=1.72in]{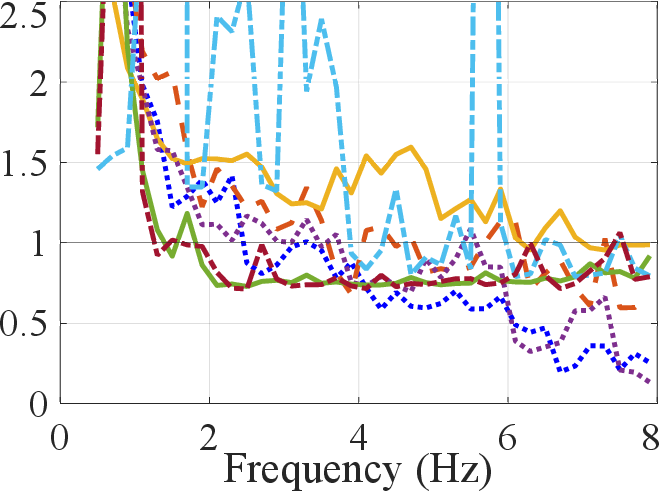}}
	\hspace{-3mm}
	\subfigure[]{
		\label{IP_CASE:c}
		\includegraphics[width=1.72in]{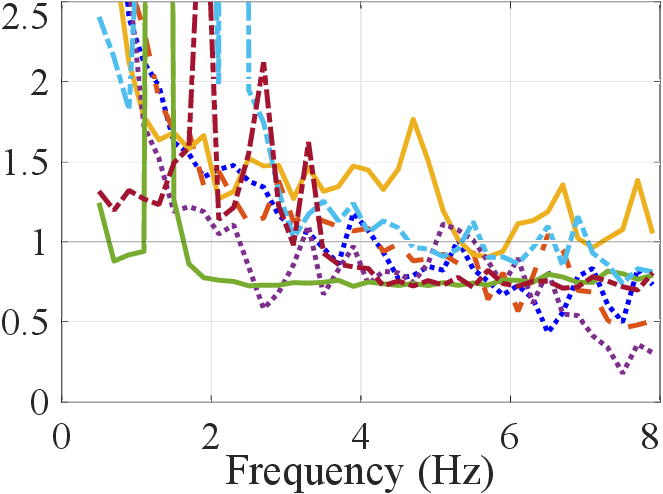}}
	\hspace{-3mm}
	\subfigure[]{
		\label{IP_CASE:d}
		\includegraphics[width=1.72in]{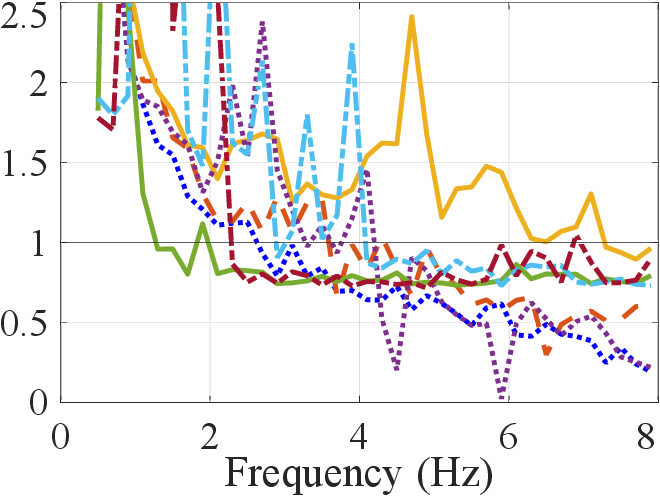}}
	\hspace{-4mm}
	\subfigure[]{
		\label{IP_TASE:a}
		\includegraphics[width=1.8in]{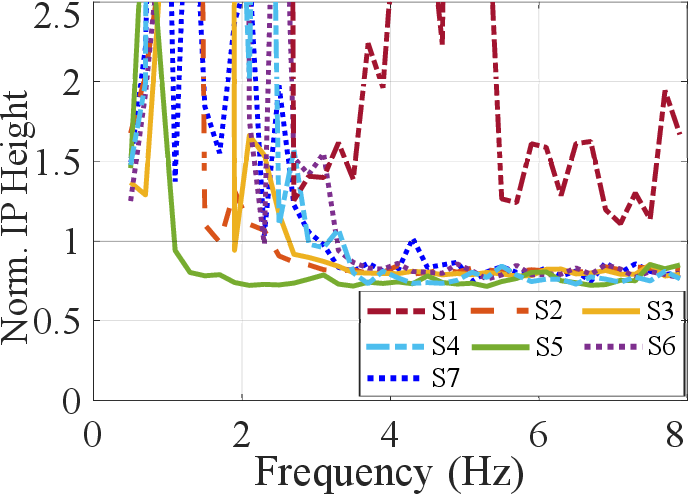}}
	\hspace{-3mm}
	\subfigure[]{
		\label{IP_TASE:b}
		\includegraphics[width=1.72in]{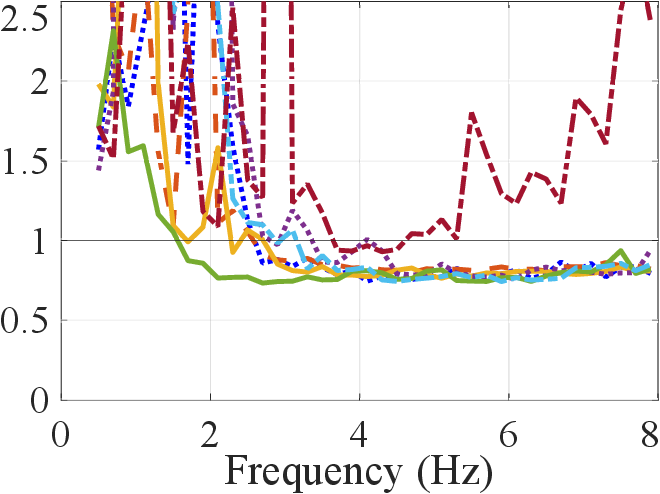}}
	\hspace{-3mm}
	\subfigure[]{
		\label{IP_TASE:c}
		\includegraphics[width=1.72in]{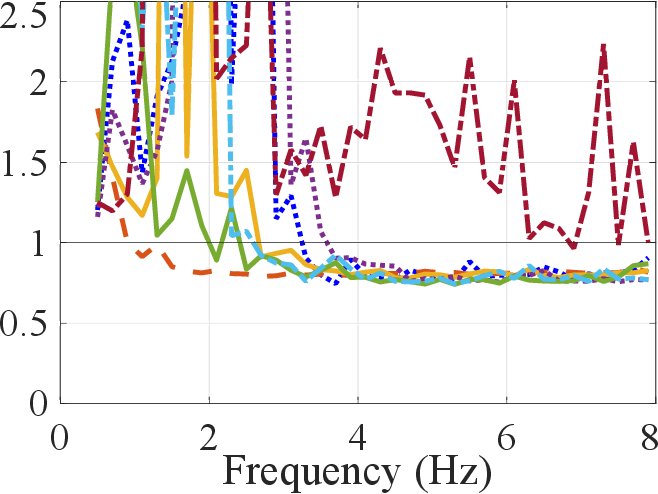}}
	\hspace{-3mm}
	\subfigure[]{
		\label{IP_TASE:d}
		\includegraphics[width=1.72in]{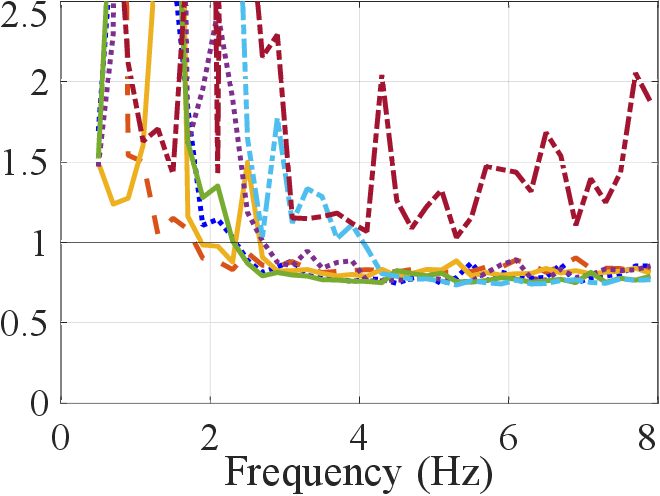}}
	\vspace{-4mm}
	\caption{Top row: Normalized stance balance task IP height-frequency curves for 7 subjects from TOI1-TOI4: (a) LE without exoskeleton; (b) HE without exoskeleton; (c) LE with exoskeleton; (d) HE with exoskeleton. Bottom row: Normalized stance welding task IP height-frequency curves for 7 subjects from TOI5-TOI8: (e) LE without exoskeleton; (f) HE without exoskeleton; (g) LE with exoskeleton; (h) HE with exoskeleton.}
	\label{IP_TASE}
	\vspace{-2mm}
\end{figure*}

The IP frequency curves offer insights into human balance control strategies. At low frequencies, the IP height greater than the COM height indicates active neuromuscular engagement and high control effort to maintain balance. As the frequency increases, the IP curve decays, suggesting a transition from active to passive control strategies, characterized by $\omega^{\text{CF}}$. The HFA slope $k^{\text{HFA}}$ of the IP curve represents the passive control mechanism dominant at higher frequencies, reflecting the neuromechanical impedance of the joints. This variation in IP height across frequencies provides insights into how muscles and joints adapt to different balance demands—offering information beyond what traditional COP or COM analysis can capture. By examining IP height at different frequencies, it becomes possible to understand how individuals adjust their balance strategies in response to various conditions, such as task complexity or elevation changes.

After extracting the IP frequency curves, statistical analysis was conducted separately for the datasets of each subject group. The behavior of the IP curve was assessed through the quantification of three key parameters, namely, $\alpha_i$, $\bs{\beta}_i$, and $\bs{\sigma}_i$, $i=t,d$, for stance and kneeling, respectively. The PSD of the COP data was also calculated to analyze the frequency content of the postural sway. The PSD quantifies the energy distribution of the COP signal across different frequency bands, allowing for the identification of dominant frequencies and their relative contributions to the overall postural control strategy. The heart rate data was incorporated into the multi-task assessment to examine the physiological responses of the subjects during the experiments. By integrating these diverse data sources, the study aims to provide a comprehensive understanding of the interplay between postural balance, exoskeleton assistance, task performance, and physiological responses in construction work. 

We also conducted simulations to analyze the impact of various neural balance controller parameter sets on the IP height frequency curve. The simulation allowed for the exploration of a wide range of scenarios and conditions that may be difficult or impossible to test experimentally. To simulate and analyze the impact of control parameters on human balance strategies, the TIP and DIP models with the LQR-based neural controller with zero initial condition to compute IP height frequency behaviors. Table~\ref{table:lumpedTIP} lists the parameter values for the models that are determined by typical human biomechanical data~\cite{Winter2009}. The simulation involved executing a series of trials to analyze the model's response under various conditions, each lasting $50$~s with data sampled at a frequency of $100$~Hz. To ensure robust statistical analysis, $30$ trials were conducted for each parameter set. 

\renewcommand{\arraystretch}{1.08}
\setlength{\tabcolsep}{0.11in}
\begin{table}[h!]
	\footnotesize
	\centering
	\caption{Model Parameters for the TIP and DIP Models}
	\vspace{-2mm}
	\label{table:lumpedTIP}
	\begin{tabular}{|c|c|c|c|c|c|} 
		\toprule[1.2pt] 
		\multicolumn{1}{|c|}{\multirow{2}{*}{Parameter}} & \multicolumn{3}{c|}{TIP} & \multicolumn{2}{c|}{DIP} \\  \cline{2-6} 
		& Link 1  & Link 2 & Link 3 & Link 2  & Link 3 \\ [0.2ex] 
		\hline
		$m_i$ ($kg$)& $6$ & $14$ & $48$ & $20$ & $42$  \\
		$l_i$ ($m$)& $0.6$ & $0.42$ & $0.7$ & $0.568$ & $0.622$ \\
		$l_{ci}$ ($m$)& $0.3$ & $0.1$ & $0.45$ & $0.284$ & $0.311$ \\
		$I_i$ ($kgm^{2}$)& $0.264$ & $0.1722$ & $0.441$ & $0.5$ & $3.5$ \\
		[0.1ex] 
		\bottomrule[1.3 pt]  
	\end{tabular}
\end{table}

\section{Results}
\label{results}

Figs.~\ref{IP_CASE:a}-\ref{IP_CASE:d} show the IP frequency curves for TOIs~1 to~4. The IP height magnitude is above the COM at low frequencies, demonstrating  highly active control effort at these frequencies. The IP curve decays with increasing frequency, indicating a reduced influence of active control. The IP curve also reaches an asymptote below the COM at high frequencies, suggesting a transition to passive control mechanisms. The spikes in the IP frequency curve indicate changes in balance strategy, which is the direct effort of an individual attempting to stabilize the gait. On examining each individual's IP curve across TOIs, several observations are obtained. In the balance tests, for instance, subject S3 generated high-frequency spikes in the IP curves with exoskeleton assistance. The observation suggests a counter-intuitive perception of instability, causing the subjects to alter their balance strategies to maintain balance. Subject S6 also showed increased variation in the IP curve with the exoskeleton. On the other hand, the presence of the exoskeleton reduced both high-frequency and high-magnitude spikes for S4. For S1, large spikes are observed for LE cases at low frequencies, while small spikes were observed for HE cases at high frequencies, indicating a shift in balance strategy on experiencing altered visual stimuli. It is evident that the interplay between HE and exoskeleton use did not uniformly impact all individuals; some subjects  (e.g., S5 and S7) maintained a consistent and stable balance strategy across all balance TOIs, regardless of elevation or exoskeleton usage.

Figs.~\ref{IP_TASE:a}-\ref{IP_TASE:d} show the IP curves generated by subjects performing quiet stance while conducting welding tasks. Examination of these curves across individual subjects reveals that focusing on tasks (TOIs~5 to~8) generally enhances postural stability. When compared to the quiet balance, most subjects in kneeling gait exhibited smoother asymptotic behavior in the IP frequency response, which suggests a stable control strategy employed during the welding tasks. This improved stability can be a result of the subjects focusing on the welding task, permitting a stable balance strategy to be instinctively chosen. An exception to this trend is observed in the case of subject S1, whose IP curve does not follow the typical characteristics expected of healthy human quiet stance. This can be attributed to the subject modifying their gait during task performance, transitioning from an upright stance to a quarter-squat position.  Thus, IP curves have the potential to discriminate between different types of gait.

\begin{figure*}[t]
	\hspace{-2mm}
	\subfigure[]{
		\label{Kneel_CASE:a}
		\includegraphics[width=1.8in]{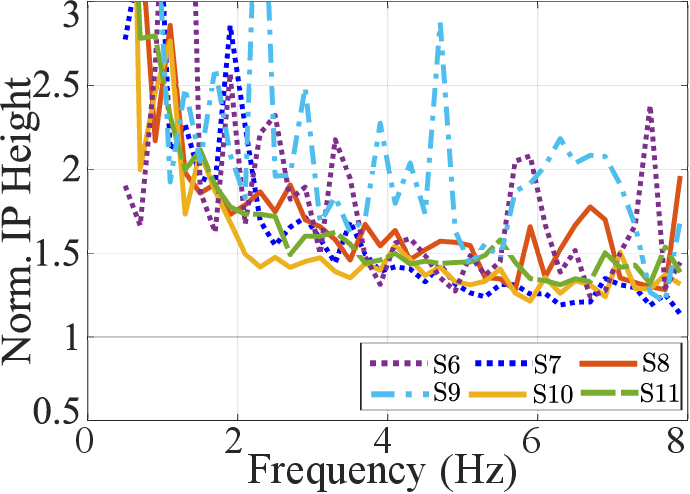}}
	\hspace{-3mm}
	\subfigure[]{
		\label{Kneel_CASE:b}
		\includegraphics[width=1.72in]{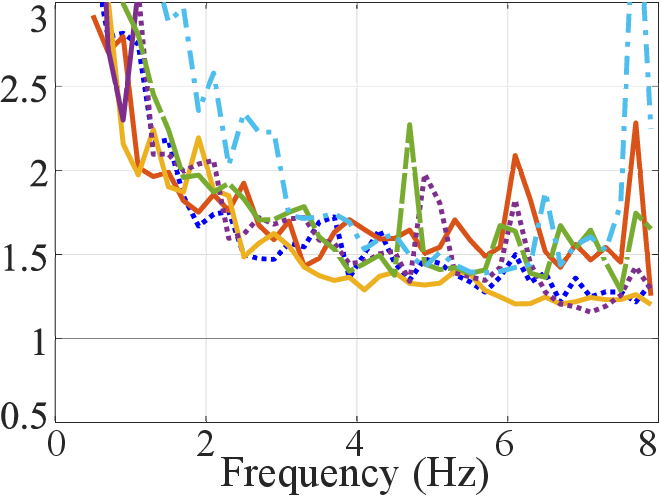}}
	\hspace{-3mm}
	\subfigure[]{
		\label{Kneel_CASE:c}
		\includegraphics[width=1.72in]{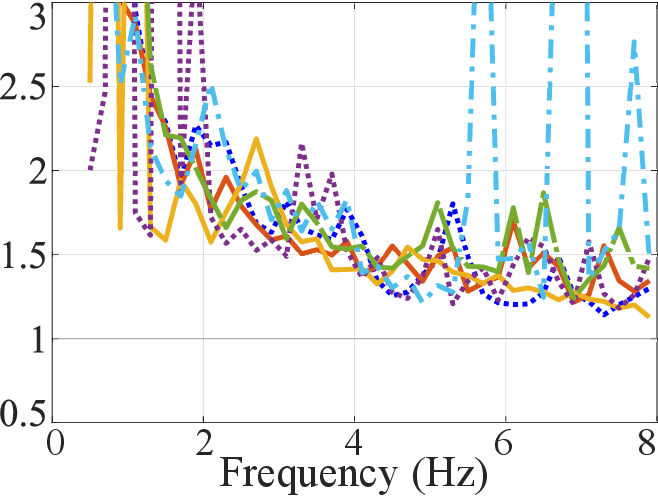}}
	\hspace{-3mm}
	\subfigure[]{
		\label{Kneel_CASE:d}
		\includegraphics[width=1.72in]{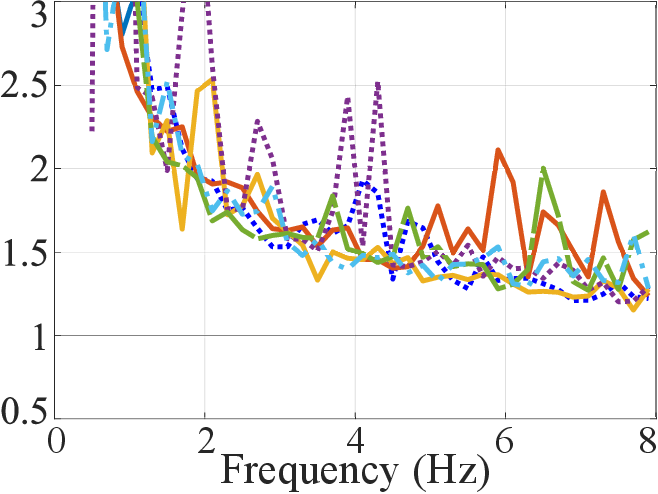}}
	\vspace{-4mm}
	\hspace{-2mm}
	\subfigure[]{
		\label{Kneeling_TASE:a}
		\includegraphics[width=1.8in]{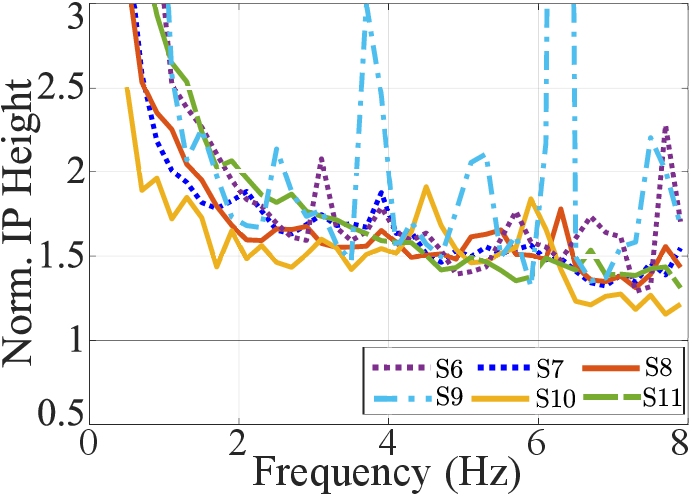}}
	\hspace{-3mm}
	\subfigure[]{
		\label{Kneeling_TASE:b}
		\includegraphics[width=1.72in]{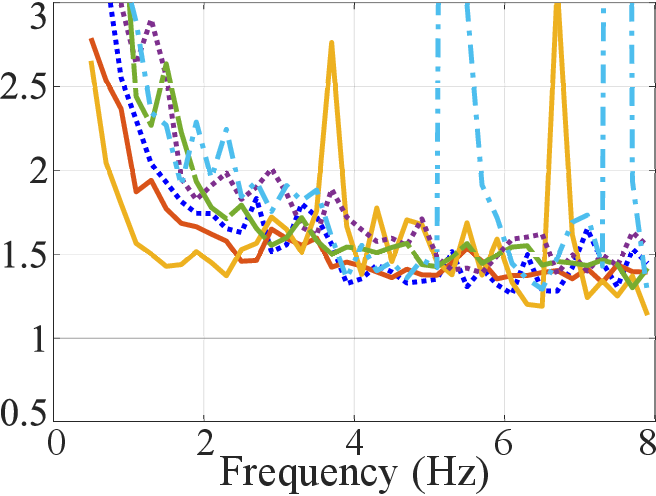}}
	\hspace{-3mm}
	\subfigure[]{
		\label{Kneeling_TASE:c}
		\includegraphics[width=1.72in]{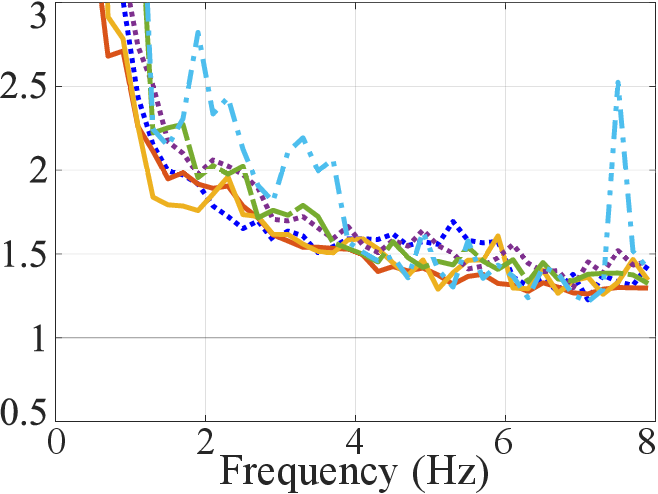}}
	\hspace{-3mm}
	\subfigure[]{
		\label{Kneeling_TASE:d}
		\includegraphics[width=1.72in]{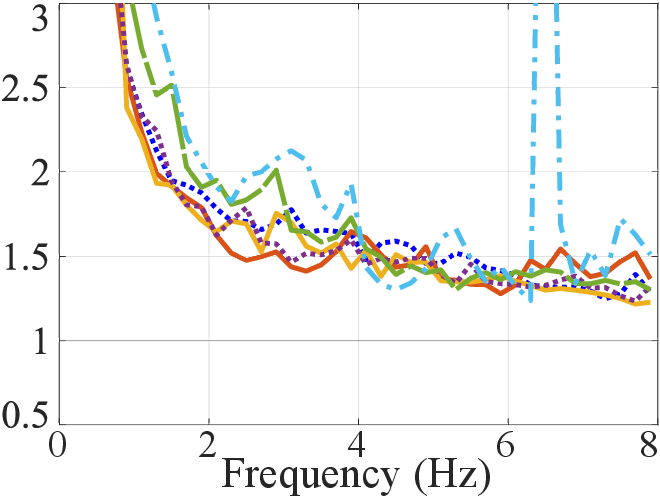}}
	\vspace{-0mm}
	\caption{Top row: Normalized kneeling balance task IP height-frequency curves for 6 subjects from TOI1-TOI4: (a) LE without exoskeleton; (b) HE without exoskeleton; (c) LE with exoskeleton; (d) HE with exoskeleton. Bottom row: Normalized Kneeling welding task IP height-frequency curves for 6 subjects from TOI5-TOI8: (a) LE without exoskeleton; (b) HE without exoskeleton; (c) LE with exoskeleton; (d) HE with exoskeleton.}
	\label{Kneeling_TASE}
	\vspace{-0mm}
\end{figure*}

Fig.~\ref{Kneeling_TASE} shows the IP height-frequency curves for the kneeling balance task and the kneeling welding task for 6 subjects. All kneeling curves are similar to the cases of quiet stance. The IP height in relation to the COM is overall higher in kneeling than in stance. The kneeling IP curves give insight into individual balance strategies in response to different testing stimuli. For the balance experiments, HE increased the magnitude of high-frequency spikes in S8 and S11 and reduced the spikes in S6. The exoskeleton assistance reduced and increased high-frequency spikes in LE balance tests for S8 and S9, respectively. When outfitted with the exoskeleton, S6, S7, and S10 show smoother IP curves during  welding task. S9's balance strategy varied greatly across all TOIs with high-frequency spikes in all cases except TOI~4. Overall, S6, S8, and S11 have smoother IP curves for TOIs~6 to 8 than those in TOIs~1 to 4.

\begin{figure*}[th!]
	\hspace{-4mm}
	\subfigure[]{
		\label{PSD:a}
		\includegraphics[width=1.85in]{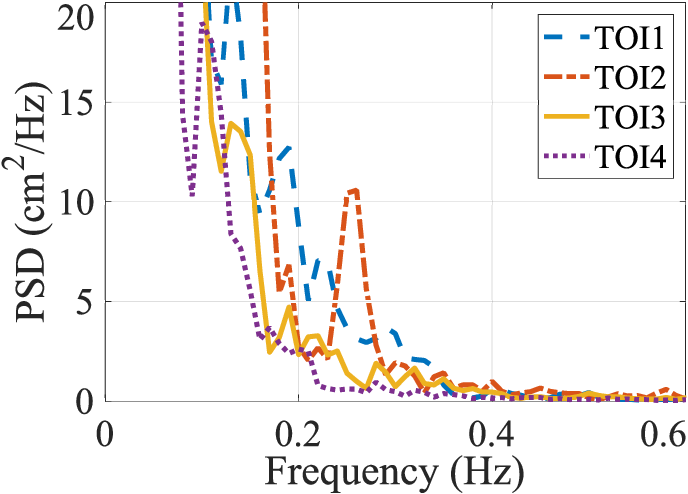}}
	\hspace{-3mm}
	\subfigure[]{
		\label{PSD:b}
		\includegraphics[width=1.7in]{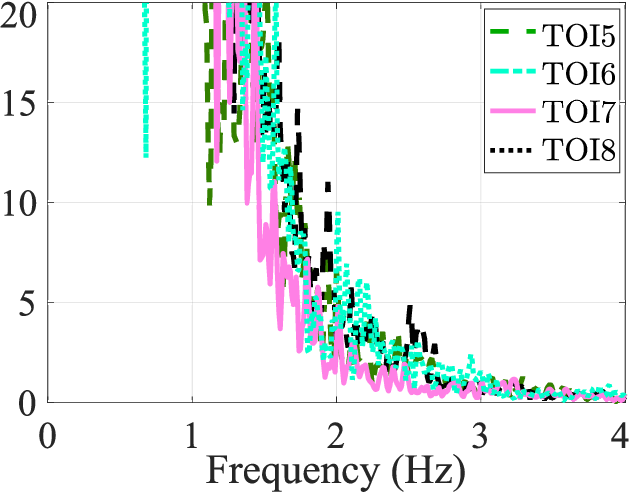}}
	\hspace{-3mm}
	\subfigure[]{
		\label{PSD:c}
		\includegraphics[width=1.7in]{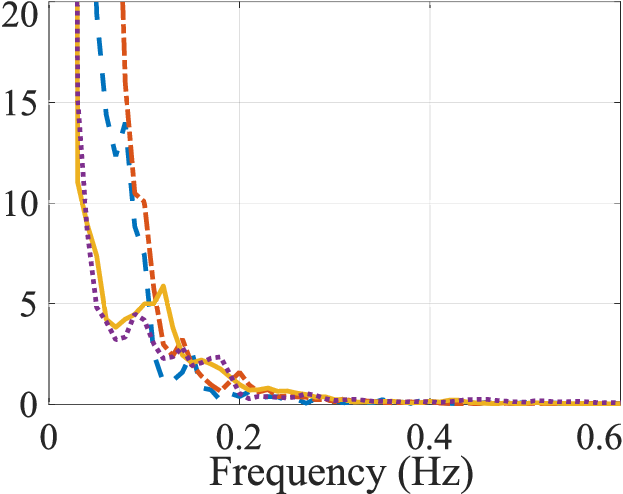}}
	\hspace{-3mm}
	\subfigure[]{
		\label{PSD:d}
		\includegraphics[width=1.7in]{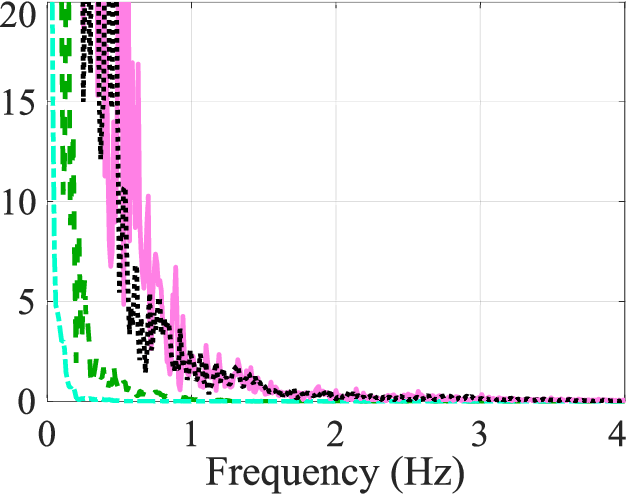}}
	\vspace{-2mm}
	\caption{Power spectral density of experimental COP results for (a) S4 under Protocol A with stance. (b) S4 under Protocol B with stance. (c) S7 under Protocol A with kneeling. (d) S7 under Protocol B with kneeling.}
	\label{PSD}
\end{figure*}

\renewcommand{\arraystretch}{1.25}
\setlength{\tabcolsep}{0.045in}
\begin{table}[ht!]
	\vspace{-0mm}
	\footnotesize
	\centering
	\caption{Mean IP Descriptor Values for all TOIs}
	\label{table:CFHFA}
	\vspace{-1mm}
	\begin{tabular}{|c|c|c|c|c|c|c|c|c|}
		\toprule[1.2pt] 
		TOI & $1$ & $2$ & $3$ & $4$ & $5$ & $6$ & $7$ & $8$ \\ \hline
		$\bar{\omega}^{\text{CF}}_t$ (Hz)   & $3.2$ & $3.6$ & $3.2$ & $3.4$ & $2.2$ & $2.7$ & $2.4$ & $2.6$ \\ 
		$|\bar{k}^{\text{HFA}}_t|$ & $0.103$ & $0.068$ & $0.117$ & $0.010$ & $0.018$ & $0.022$ & $0.012$ & $0.018$ \\  
		$\bar{\omega}^{\text{CF}}_d$  (Hz)   & $3.9$  & $3.9$ & $4.1$ & $4.1$ & $5$ & $3.9$ & $5$ & $4.2$ \\ 
		$|\bar{k}^{\text{HFA}}_d|$ & $0.023$ & $0.014$ & $0.014$ & $0.043$ & $0.018$ & $0.052$ & $0.047$ & $0.033$ \\ 
		[0.1ex] 
		\bottomrule[1.3 pt]  
	\end{tabular}
	\vspace{-0mm}
\end{table}

The one-way analysis of variance (ANOVA) was conducted to analyze the differences in $\omega_i^{\text{CF}}$ and $k_i^{\text{HFA}}$ across all TOIs. Table~\ref{table:CFHFA} presents the resulting mean values and statistical significance of these parameters. Upon the introduction of the exoskeleton, there was a discernible decrease in the mean CF $\bar{\omega}_t^{\mathrm{CF}}$ during the balance experiments. This suggests a shift in the balance control strategy employed by the subjects. The value of $\bar{\omega}_t^{\mathrm{CF}}$ was also found to increase with the HE visuals. For the welding task scenario, overall $\bar{\omega}_t^{\mathrm{CF}}$ is lower than that of the balance scenario, indicating the introduction of the task reduced balance control effort. While the combination of the exoskeleton and HE visuals lowered $\bar{\omega}_t^{\text{CF}}$ in the balancing experiments, performing welding tasks increased its value. The mean slope of the HFA $|\bar{k}_t^{\text{HFA}}|$ had a higher magnitude for the balance experiments than the welding task experiments. The condition of HE without exoskeleton had the lowest magnitude of $|\bar{k}_t^{\text{HFA}}|$, while LE with exoskeleton support had the highest $|\bar{k}_t^{\text{HFA}}|$ in balancing experiments. The vice-versa was true for the welding task performance experiments. The kneeling IP descriptors are also presented in Table.~\ref{table:CFHFA}. The value of $\bar{\omega}^{\text{CF}}_d$ does not vary with visual stimuli. It however increases with the exoskeleton assistance and welding task. The task performance increases the magnitude of $|\bar{k}_d^{\text{HFA}}|$ for HE without exoskeleton assistance and LE with exoskeleton assistance. For LE without exoskeleton support and HE with exoskeleton support, the values of $|\bar{k}_d^{\text{HFA}}|$ however decreased.

Fig.~\ref{PSD} shows the PSD curves under various test conditions. The figures compare the PSD results across the balance and welding tasks during stance and kneeling gaits. By examining the PSD plots, we can concisely analyze the postural sway behavior. For example, low postural sway is observed in welding task TOIs (asymptote begins around $0.4$-$0.6$~Hz) compared to the balance TOIs (asymptote begins around $3$-$4$~Hz), regardless of the gait performed. Stance overall has higher postural sway than kneeling gait as showcased by the PSD frequency curves located at high frequencies. High-elevation visual feedback caused more postural sway during stance than kneeling gait as seen by the higher area under the curve. While the exoskeleton reduced the sway in most TOIs, it allowed for increased sway in TOIs 7 and 8 during kneeling. As shown in Figs.~\ref{IP_TASE} and~\ref{Kneeling_TASE}, the IP metric complements these observations by providing detailed physiological and joint-level insights. 

\begin{figure}[h!]
    \centering
    \includegraphics[width = 3.2in]{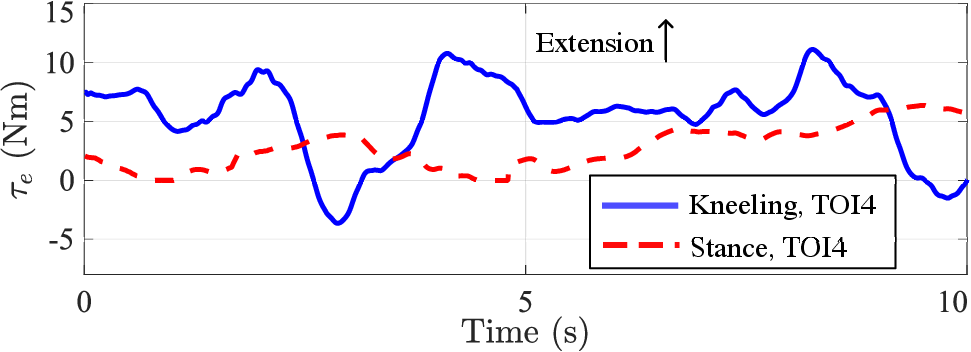}
    \vspace{-1mm}
    \caption{An example of knee exoskeleton output torque $\tau_e$ for both stance and kneeling gaits in TOI~4.}
    \label{fig:tau}
\end{figure}

\begin{figure}[h!]
	\centering
	\includegraphics[width = 3.15in]{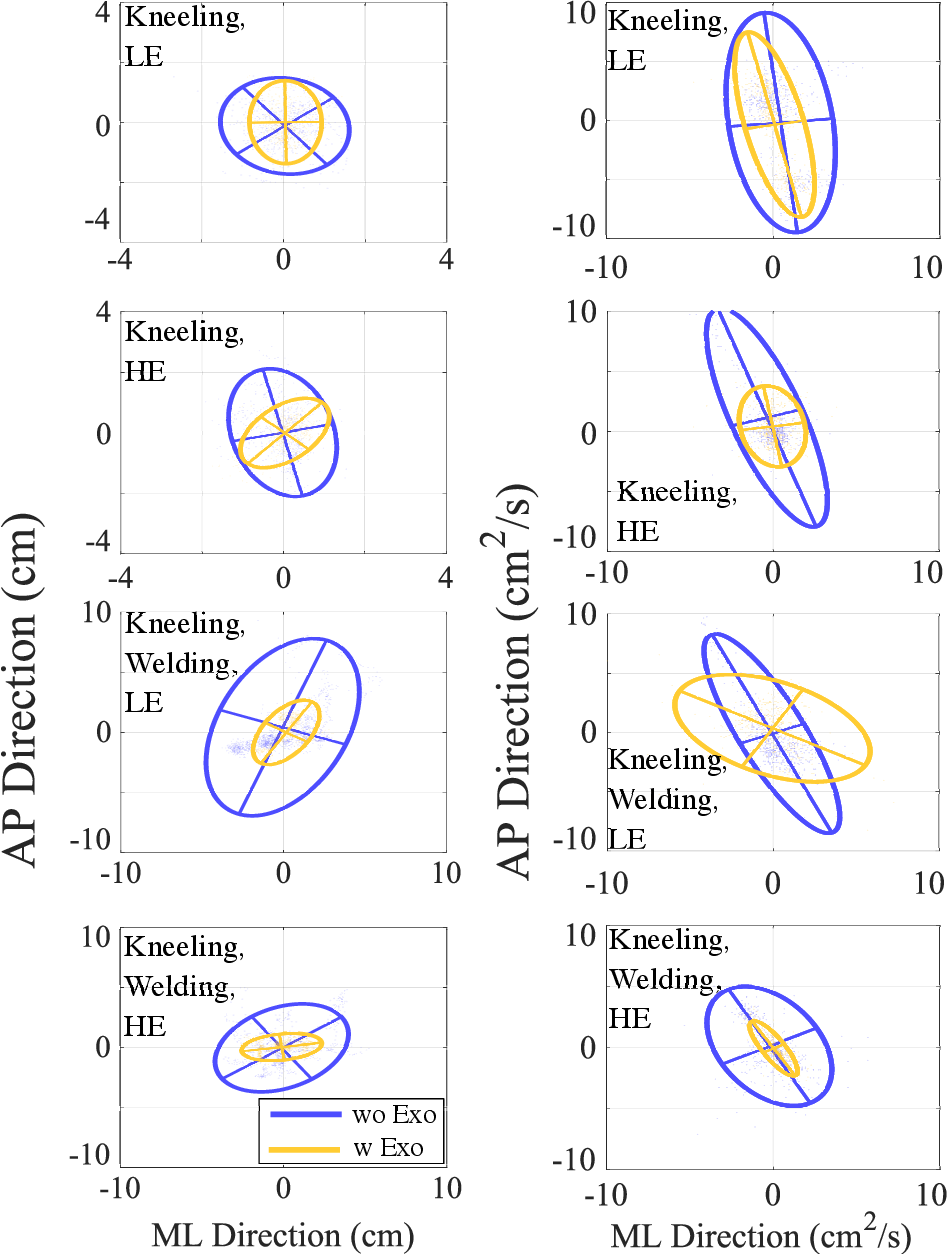}
	\vspace{-1mm}
	\caption{$A^{\text{Sway}}_d$ (left) and $A^{\text{Acc}}_d$ (right) for a representative subject (S7) during kneeling experiments under different TOIs.}
	\label{fig:SwayVSAcc}
	\vspace{-0mm}
\end{figure}

Fig.~\ref{fig:tau} shows an example of assistive torque $\tau_e$ by the exoskeleton during a 10-s interval for both stance and kneeling gaits in TOI~4 from one subject. The stiffness-based controller during quiet stance generated extension torques and the PD controller for kneeling demonstrated dynamic torque range. These torque outputs helped maintain the knee joint's desired position and stability. The effectiveness of the exoskeleton assistance in improving balance can be further observed through metrics such as the $A_d^{\text{Sway}}$ and $A_d^{\text{Acc}}$, as shown in Fig.~\ref{fig:SwayVSAcc}. The reduced ellipse areas in the presence of the exoskeleton, particularly during HE conditions and welding tasks, indicate that the device contributes to better postural stability and reduced body sway.  

\begin{table*}[h!]
	\renewcommand{\arraystretch}{1.25}
	\setlength{\tabcolsep}{0.067in}
	\centering
	\footnotesize
	\centering
	\caption{Mean Sway Area ($\text{cm}^2$) and Sway Acceleration Area ($\text{cm}^2$/s)}
	\vspace{-1mm}
	\label{table:Ellipse}
	\begin{tabular}{|c|c|c|c|c|c|c|c|c|}
		\toprule[1.2pt] 
		TOI               & $1$  & $2$ & $3$ & $4$ & $5$ & $6$ & $7$ & $8$  \\ \hline
		$A^{\text{Acc}}_t$  & $30.1 \pm 21.6$ & $38.7 \pm 24.9$ & $\boldsymbol{17.2 \pm 16.4}$ & $\boldsymbol{14.9 \pm 9.9}$ & $248.2 \pm 354.8$ & $371.2 \pm 562.5$ & $\boldsymbol{275.8 \pm 312.2}$ & $\boldsymbol{243.4 \pm 259.6}$ \\ 
		$A^{\text{Sway}}_t$ & $43.3 \pm 37.4$ & $32.4 \pm 21.2$ & $\boldsymbol{15.3 \pm 16.7}$ & $\boldsymbol{31.9 \pm 28.6}$ & $109.3 \pm 92.5$ & $101.8 \pm 68.2$ & $134.5 \pm 107.9$ & $\boldsymbol{101.4 \pm 77.4}$ \\ 
		$A^{\text{Acc}}_d$  & $19.9 \pm 21.7$ & $9.4 \pm 2.7$ & $\boldsymbol{6.1 \pm 6.0}$ & $\boldsymbol{10.9 \pm 7.7}$ & $25.6 \pm 19.8$ & $35.4 \pm 29.2$ & $\boldsymbol{20.5 \pm 14.7}$ & $\boldsymbol{14.5 \pm 4.9}$ \\ 
		$A^{\text{Sway}}_d$ & $2.6 \pm 2.8$ & $2.9 \pm 3.4$ & $\boldsymbol{1.5 \pm 1.3}$ & $\boldsymbol{1.8 \pm 1.2}$ & $14.4 \pm 9.8$ & $24.7 \pm 23.7$ & $18.4 \pm 20.3$ & $\boldsymbol{21.1 \pm 15.5}$ \\ 
		[0.1ex] 
		\bottomrule[1.3 pt]  
	\end{tabular}
	\vspace{-2mm}
\end{table*}

Fig.~\ref{fig:StanceVSKneel} further compares the sway acceleration ellipses and the postural sway ellipses for  TOIs~2 and~4 experiments. The relatively larger ellipse areas for the kneeling gait compared to the stance gait, particularly under TOI~2, suggest that kneeling is inherently less stable than stance. In the presence of higher cognitive load and visual disturbances, the exoskeleton provides significant support and stabilization during kneeling tasks, as evidenced by the reduced ellipse size compared to TOI~2. Table~\ref{table:Ellipse} lists the mean and standard deviation of $A_t^{\text{Acc}}$ and $A_t^{\text{Sway}}$ as quantitative assessments of postural stability. For stance, there is a notable decrease in average $A_t^{\text{Sway}}$ from TOI~1 and~2, signaling that subjects reduced their sway in response to visually induced stress. However, an increase of  $A_t^{\text{Acc}}$ suggests that the subjects responded with significant dynamic control to maintain this constrained posture. The stabilizing effect of the exoskeleton was evident in TOI~3, where reduced $A_t^{\text{Sway}}$ and $A_t^{\text{Acc}}$ correspond to a relaxed stance, while in TOI~4, with high $A_t^{\text{Sway}}$ but low $A_t^{\text{Acc}}$ values, indicates deliberate and controlled movements. Welding tasks, particularly TOI~6, show the largest $A_t^{\text{Acc}}$ and $A_t^{\text{Sway}}$, indicating that engagement in the task necessitates broader movements. For the kneeling condition, the overall $A_d^{\text{Acc}}$ and $A_d^{\text{Sway}}$ are smaller compared to the quiet stance condition, suggesting that the kneeling posture inherently constrains motion. 

\setcounter{figure}{11}
\begin{figure}[h!]
    \centering
    \includegraphics[width = 3.2in]{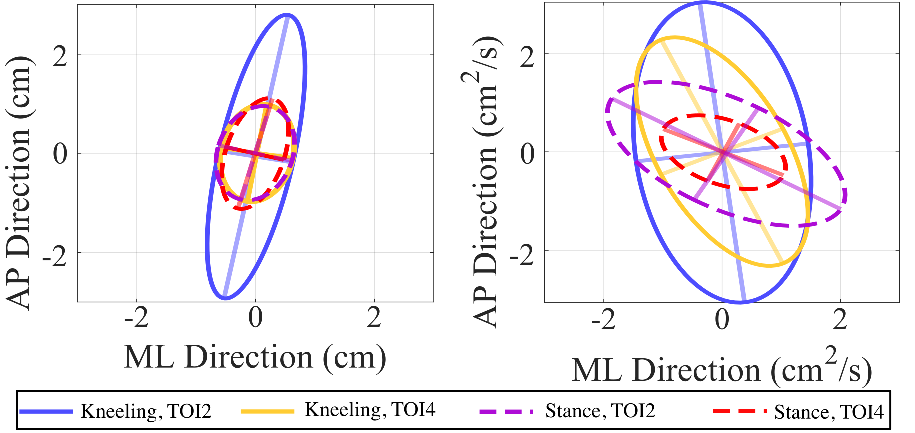}
    \caption{Comparison of $A^{\text{Sway}}_i$ (left) and $A^{\text{Acc}}_i$ (right) between kneeling and stance gaits under TOIs~2 and~4.}
    \label{fig:StanceVSKneel}
\end{figure}

\setcounter{figure}{12}
\begin{figure*}[h!]
	\centering
	\subfigure[]{
		\includegraphics[width=3.55in]{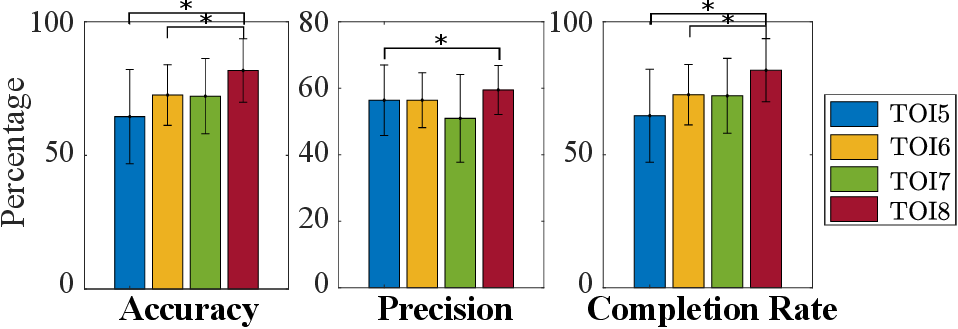}
	}
	\subfigure[]{
		\includegraphics[width=3.2in]{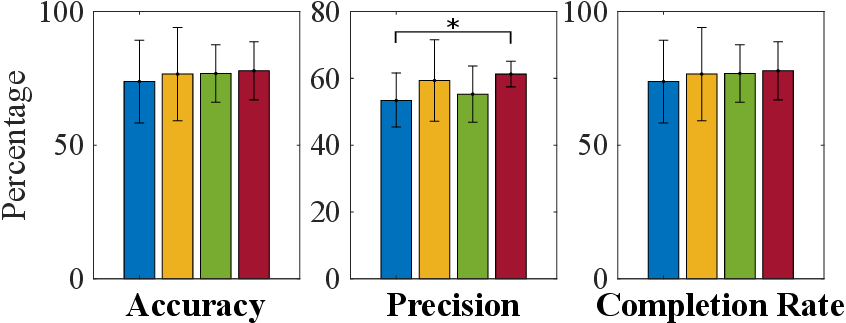}
	}
	\vspace{-3mm}
	\caption{Performance metrics for the virtual welding task across four TOIs for (a) quiet stance and (b) kneeling gaits. Error bars represent $\pm 1$ standard error. Asterisks indicate significant differences between the two groups ($p < 0.05$). }
	\label{fig:welding_performance}
	\vspace{-0mm}
\end{figure*}


Fig.~\ref{fig:welding_performance} shows the task performance through three metrics for the welding task across TOI~5 to~8. The ANOVA analysis revealed significant differences across TOIs for performance metrics. For quiet stance, it is observed that the incorporation of the exoskeleton in TOIs~7 and~8 correlates with improved accuracy and completion rate. Specifically, accuracy and completion rate were significantly higher in TOI~8 compared to TOI~5 ($p < 0.05$) and this suggests that the exoskeletons improve performance at high elevations. The completion rate in HE conditions (i.e., TOI~8) is particularly enhanced with exoskeleton assistance, whereas precision appears to be less influenced by the elevation or the presence of the exoskeleton. The kneeling gait exhibits a similar trend in task performance metrics. Precision was significantly higher in TOI~8 compared to TOI~5 ($p < 0.05$). The use of the exoskeleton in TOIs~7 and~8 in kneeling gaits also leads to increased accuracy and completion rate compared to TOIs~5 and~6 without exoskeleton assistance. This finding suggests that the exoskeleton's benefits extend beyond a quiet stance and effectively support task execution in kneeling postures as well. However, it is worth noting that the overall accuracy and completion rate values for kneeling are slightly lower than those for quiet stance across all TOIs. This difference may be attributed to the inherent challenges associated with maintaining balance and performing tasks in kneeling posture.


Fig.~\ref{fig:HR} shows the heart rate data collected during the welding tasks for both gaits. During quiet stance, the heart rate increase from LE to HE suggests a progressive elevation in stress levels, with the highest median heart rate recorded during the HE tasks, regardless of exoskeleton usage. This incremental pattern reflects the physiological demands imposed by elevation changes and the complexity of tasks performed under such conditions. Statistical analysis using ANOVA revealed that the differences in heart rates between the conditions were not statistically significant ($p > 0.05$), indicating that while heart rates increased from LE to HE, the variability among subjects may have masked any significant group-level effects. Nevertheless, the trends observed, particularly the reduction in heart rate with exoskeleton assistance, align with our hypothesis that the exoskeleton alleviates physical stress during tasks at both LE and HE. During kneeling gaits, although the application of the exoskeleton reduced human effort and lowered the heart rate, the heart rate was generally higher compared to a quiet stance. The observed trend of lower heart rates with exoskeleton usage suggests potential physiological benefits in reducing stress, even if the differences did not reach statistical significance. Table~\ref{table:HR} lists the average heart rates from some subjects for TOI~5 to~8. The data reveal minimal fluctuations in heart rate among most subjects across different TOIs, suggesting that the welding task may serve as a focal stimulus, concentrating the participants' attention and thereby stabilizing physiological responses \cite{habibnezhad2020neurophysiological}. Subjects S2, S3, and S5 demonstrated notable variations in average heart rate upon the integration of the exoskeleton into the task during stance gait. Similar trends are observed during kneeling gaits, where most subjects experienced a reduction in heart rates with exoskeleton assistance while performing welding tasks.

\setcounter{figure}{13}
\begin{figure}[h!]
	\centering
	\subfigure[]{
		\includegraphics[width=1.6in]{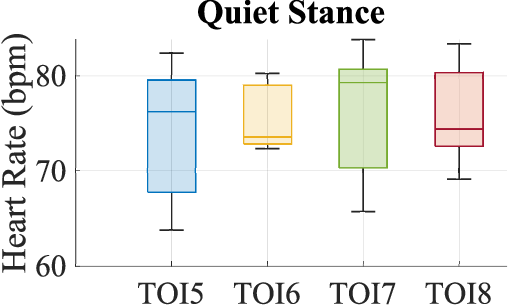}
		\label{fig:stance_HR}}
	\subfigure[]{
		\includegraphics[width=1.5in]{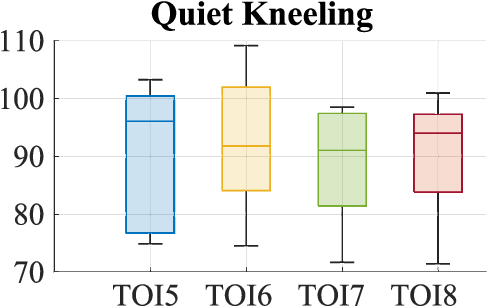}
		\label{fig:kneeling_HR}}
	\caption{Heart rate comparison for the virtual welding task across four TOIs during (a) quiet stance and (b) kneeling gaits. }
	\label{fig:HR}
\end{figure}

\setcounter{table}{3}
\renewcommand{\arraystretch}{1.3}
\setlength{\tabcolsep}{0.07in}
\begin{table}[ht!]
	\footnotesize
		\centering
		\caption{Mean Heart Rate Data (in BPM)}
\vspace{-2mm}
		\label{table:HR}
		\begin{tabular}{|c|c|c|c|c|} 
		\toprule[1.2pt] 
			\multicolumn{5}{|c|}{Quiet Stance/Kneeling} \\ 	\hline
			Subject & TOI5 & TOI6 & TOI7 & TOI8 \\ [0.2ex] 
			\hline

            S1 / S8 & $79.4/100.4$ & $79.5/87.0$ & $79.7/98.5$ & $83.4/97.3$ \\
			S2 / S9 & $66.0/103.3$ & $65.7/96.6$ & $73.0/97.5$ & $74.4/92.7$ \\
			S3 / S10 & $82.4/76.8$ & $81.1/84.1$ & $73.6/81.4$ & $72.6/83.8$ \\
			S4 / S11 & $79.6/74.9$ & $83.8/74.5$ & $80.3/71.7$ & $78.2/71.4$ \\
			S5  & $63.8/-$ & $69.6/-$ & $72.4/-$ & $72.8/-$ \\
			S6  & $73.0/95.8$ & $72.5/109.2$ & $72.8/87.0$ & $69.1/95.4$ \\
			S7 & $76.3/96.3$ & $79.3/102.0$ & $77.0/95.2$ & $81.1/101.0$ \\
		[0.1ex] 
        \bottomrule[1.3 pt]  
		\end{tabular}
\vspace{-3mm}
\end{table}

Table~\ref{table:questionnaire_results} summarizes the results from the subjects' questionnaire feedback. The data indicates a wide range of susceptibility to fear and acrophobia across all subjects for both stance and kneeling gaits. In stance experiments, subjects S2 and S7 showed the highest levels of susceptibility to fear and acrophobia, whereas S1 displayed the lowest. For kneeling tests, S11 exhibited the highest susceptibility to fear and S9 had the lowest. Cognitive loading was reported to be higher during tasks at HE for most subjects than that at LE in both stance and kneeling experiments. However, subjects S1 and S2 in stance and S6 in kneeling experiments reported high cognitive load with the exoskeletons at HE. When considering task performance, subjects generally focused best on the addition of the exoskeletons, while their focus was compromised without the exoskeletons. This trend was observed in both gaits experiments, suggesting that the exoskeleton might offer a psychological benefit in high-stress situations, regardless of the working posture.

\setcounter{table}{4}
\renewcommand{\arraystretch}{1.32}
\setlength{\tabcolsep}{0.052in}
\begin{table*}[t!]
	\centering
	\caption{Subject Questionnaire Results}
	\label{table:questionnaire_results}
	\vspace{-2mm}
	\footnotesize
	\begin{tabular}{|c|c|c|c|c|c|c|c|c|c|c|c|c|}
		\toprule[1.2pt] 
		\multicolumn{2}{|c|}{Subject ID} &  S1 & S2 & S3  & S4  & S5  &  S6 & S7  & S8 & S9 & S10 & S11   \\ \hline 
		\multicolumn{1}{|c|}{\multirow{1}{*}{JGF Quest.}} &  Suscept. fear &  $\boldsymbol{63}$   & $\boldsymbol{123}$ & $93$ & $98$ & $65$ & $112$ & $\boldsymbol{125}$ & $85$ & $\boldsymbol{61}$ & $89$ & $\boldsymbol{132}$ \\  \hline
		\multicolumn{1}{|c|}{\multirow{2}{*}{CA Quest.}} & Fear Heights & $22$ & $56$ & $40$ & $40$ & $27$ & $53$ & $54$ & $33$ & $48$ & $45$ & $60$ \\ \cline{2-13} 
		& Avoid. HE & $22$ & $36$ & $26$ & $28$ & $23$ & $36$ & $34$ & $27$ & $49$ & $36$ & $37$ \\ \hline
		\multicolumn{1}{|c|}{\multirow{2}{*}{Cog. Load}} & Stressful  & \textbf{HE+Exo}     & \textbf{HE+Exo} & HE     & HE & HE     & HE/\textbf{HE+Exo}     & HE & HE& HE& HE& HE\\ \cline{2-13} 
		& Relaxed & LE & LE+Exo & LE+Exo & LE     & LE     & LE/LE+Exo      & LE+Exo     & LE+Exo& LE+Exo& LE+Exo& LE\\ \hline 
		\multicolumn{1}{|c|}{\multirow{2}{*}{Task Perf.}} & Most Focus   & HE & HE+Exo & HE+Exo & LE+Exo     & HE+Exo & LE/HE+Exo & HE+Exo & LE+Exo& LE+Exo& LE+Exo& LE+Exo    \\ \cline{2-13} 
		& Least Focus            & LE+Exo     & LE     & HE     & HE & HE     & LE+Exo/LE     & LE & HE& HE& HE& HE\\ [0.1ex] 
		\bottomrule[1.3 pt]  
	\end{tabular}
\end{table*}

All subjects reported feeling tense during HE scenarios and calm during low elevation scenarios, with varied responses to the presence of the exoskeleton. However, it should be noted that the subjects' perception of focus and relaxation did not always equate to good postural balance and balancing choices. Stressful situations, such as focusing on welding, can lead to instinctive balance strategies being chosen, resulting in better balance. Perturbations under stressful situations can cause posture to rapidly become unstable. In the stance experiments, subjects S1 and S2 expressed an increase in stability and balance when utilizing the exoskeletons. In contrast, S7 experienced a restriction in mobility due to the  exoskeletons. Meanwhile, S5 noticed a marginal benefit, and S6 found that the exoskeletons helped maintain position during the task. The responses to height exposure were equally varied. Most subjects consistently described the HE environment as highly realistic and fear-inducing. Interestingly, all participants in the kneeling experiments reported enhanced stability and reduced effort in maintaining balance when using the exoskeletons. 

\begin{figure}[h!]
    \centering
    \includegraphics[width=3in]{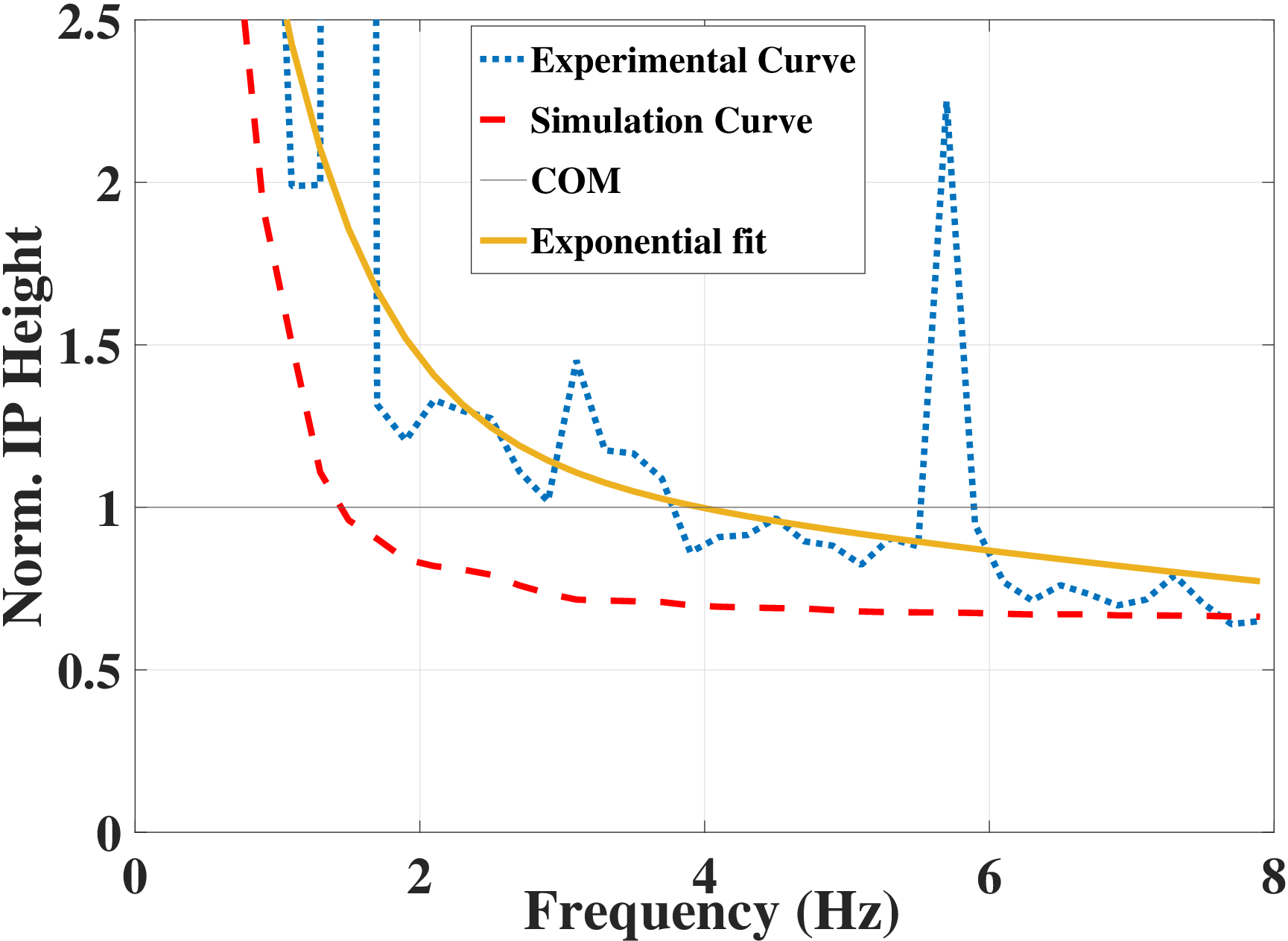}
    \vspace{-2mm}
    \caption{Comparison among the experiment, simulated, and exponentially fitted IP curves for both gaits. }
        \label{fig:exp}
\end{figure}

The best-fit parameters were derived by varying the parameter sets to obtain frequency behavior matching the observed experiments during quiet stance across various TOIs. Fig.~\ref{fig:exp} shows an example of exponential fitting the stance and kneeling IP curves. Table~\ref{LQR1} lists the best-fit parameter sets that quantify the favored balance strategy corresponding to each TOI. The parameter values were obtained by minimizing the sum of the  angle $q(t)$ errors from all subject experiments. 

Examining the variations in ${\alpha}$ values reveals a characteristic for minimal control effort. A high ${\alpha}$ magnitude signifies a minimal control effort and is generally chosen by humans during stance and this tendency however differs across TOIs. For the balancing experiments, all cases had a high magnitude of ${\alpha}$, with TOI~2 utilizing the lowest control effort. The same results were observed from the task performance scenarios TOIs~5, 6, and 8, while TOI~7 had a low magnitude of ${\alpha}=10$. The low value of ${\alpha}$ implies that a non-minimal effort control strategy was unexpectedly chosen to balance. This anomaly potentially suggests that the subjects could have altered their balance strategies to adapt to the exoskeleton in the presence of non-threatening visuals, rather than passively rely on the extra support it provides. For the rest of the welding cases, ${\alpha}$ value (i.e., $10^{10}$) is high, which implies that the welder's main focus on precision during the welding task caused them to use a minimal effort balance strategy.

\renewcommand{\arraystretch}{1.1}
\setlength{\tabcolsep}{0.075in}
\setcounter{table}{5}
\begin{table}[ht!]
	\vspace{-0mm}
	\centering
	\caption{Best-fit LQR controller parameters for quiet stance TOI}
	\label{LQR1}
	\vspace{-2mm}
	\footnotesize
	\begin{tabular}{|c|c|c|c|c|c|c|c|c|} 
		\toprule[1.2pt] 
		TOI & Case & $\alpha$ & $\beta_{1}$ & $\beta_{2}$ & $\beta_{3}$ & $\sigma_{1}$&$\sigma_{2}$&$\sigma_{3}$\\ [0.5ex] 
		\hline
		1 & LE & $10^{6}$ & $0.2$ & $0.1$ & $0.3$ & $1$ & $1$ & $1$\\ 
		2 & HE & $10^{10}$ & $0.3$ & $0.1$ & $33.3$ & $1$ & $1$ & $1$\\
		3 &  LE+Exo & $10^{6}$ & $0.2$ & $0.1$ & $0.3$ & $0.7$ & $0.3$ & $1$\\
		4 & HE+Exo & $10^{6}$ & $0.2$ & $0.1$ & $0.3$ & $0.7$ & $0.3$ & $1$ \\
		\hline
		5 & LE Weld& $10^{10}$ & $0.2$ & $0.1$ & $0.3$ & $1$ & $1$ & $1$\\ 
		6 & HE Weld& $10^{10}$ & $0.2$ & $0.1$ & $0.3$ & $1$ & $0.7$ & $0.3$\\
		7 & LE+Exo Weld & $10$ & $0.3$ & $0.1$ & $33.3$ & $1$ & $1$ & $1$\\
		8 & HE+Exo Weld & $10^{10}$ & $0.2$ & $0.1$ & $0.3$ & $1$ & $1$ & $1$ \\
		[0.1ex] 
		\bottomrule[1.3 pt]  
	\end{tabular}
	\vspace{-0mm}
\end{table}

The LQR parameter $\bs{\beta}$ provides insights into joint-specific balance strategies and reveals how control efforts are distributed across different joints. Large $\beta_i$ value ($i=1,2,3$) implies a minimal control effort to the corresponding joint. Across all TOIs, it was found that $\beta_{2}<\beta_{1}<\beta_{3}$, which suggests employing hip strategy is penalized compared to employing ankle-knee strategies. TOIs~2 and~7 have higher $\beta_3$ values, with subjects favoring the ankle-knee strategy heavily over the hip strategy to balance. The preference for the ankle-knee strategy in TOI~2 can be the consequence of the destabilizing visuals, with the subjects attempting to keep their head, the visual source of the body, stable and upright. For possible similar reasons, TOI~6 shows relatively higher noise values at the ankle ($\sigma_1$) and knee joints ($\sigma_2$) than at the hip joint ($\sigma_3$). For TOI~7, the high-effort control strategy suggests that the ankle-knee strategy chosen, while intentional, required much effort to maintain balance. The neuromechanical noise at the knee joint is lower with the presence of the exoskeleton for the stance case (TOIs~3 and~4). On introducing the VR task scenario, the neuromechanical noise experienced is similar at all joints with the exoskeleton. A similar balance strategy is used for both TOI 3 and 4, thus exhibiting reduced the influence of visual stimuli on balance with the exoskeleton in the absence of task performance.

\section{Discussion}
\label{dis}

The balance strategies discerned from IP curve analyses suggest active neuromuscular engagement at lower frequencies and a transition to passive control strategies at higher frequencies, modulated by the used knee exoskeleton. The decrease in mean CF with exoskeleton application implies an adaptive response in the human balance control system, possibly reflecting augmented stability. The consistency in the mean HFA slope across TOIs with exoskeleton intervention indicates uniform neuromechanical impedance on the joints, with the high slope interpreted as an increment in neuromechanical noise and a strategic shift in balance at high frequencies. The altered strategy can be a consequence of the individuals' effort to recalibrate their balance strategy to accommodate the exoskeleton during task execution (e.g., TOI~7), as opposed to being passively supported in a static posture (TOI~8). The results from Tables~\ref{table:CFHFA} align with the notion in~\cite{sreenivasan2023neural} that while exoskeletons can enhance postural control, their impact on the neuromechanical behavior of the body is predictable and can be characterized by specific frequency-domain features of the IP curve. These insights are pivotal for optimizing the design of exoskeletons to harmonize with the body's natural balance strategies. By observing the parameter $\alpha$, it is also found that kneeling balance control requires much higher effort than during stance to maintain postural stability. 

As shown in Fig.~\ref{IP_TASE}, the introduction of the welding task increased the presence of low-frequency spikes in the rest of the subjects' IP curves. Low-frequency spikes reflect proactive intentional efforts to change the balance strategy rather than reactionary attempts to stabilize stance in response to environmental perturbations. While low-frequency spikes were reduced with exoskeleton assistance for S5, they increased for S6 for both balance and task performance TOIs. The {color{blue}subject-specific} insights drawn from the IP analysis are valuable in identifying subjects who may be susceptible to destabilization due to specific task scenarios, thereby informing necessary measures to mitigate the risk of injury.

Task performance assessments reveal the exoskeleton's potential to alleviate the cognitive and physical loads on the subject in the presence of visual disturbances in construction. The comparative analysis between quiet stance and kneeling emphasizes the need for targeted interventions and training programs addressing specific challenges associated with different working postures. The increase in heart rate during kneeling can be attributed to greater muscle activation, reduced venous return, and psychological stress \cite{gallagher2014influence,gao2011effects}. The combination of physiological metrics permits a distinction between physical and cognitive stressors, highlighting the complexity of the exoskeleton's effectiveness in high-stress tasks. The consistency in precision values suggests that the exoskeleton primarily aid in postural stability and task completion rather than fine motor control.

The IP method provides a comprehensive assessment of the worker's postural control mechanisms. This approach is particularly valuable in the context of construction work, where workers are exposed to a wide range of environmental conditions, such as elevated worksite and unstable floor surfaces. By analyzing the frequency characteristics of the IP heights, researchers and safety professionals can gain insights into how workers adapt their balance strategies in response to different work scenarios. The knee joint strategies play a crucial and non-trivial role in maintaining balance, highlighting the importance of considering knee joint control when designing balance-assistive devices or developing balance training programs for construction workers. In contrast to the previous work in~\cite{ChenCASE2021}, the current study employs a more comprehensive approach to capture the complexity of balance control in real-world construction scenarios. The research work in~\cite{ChenCASE2021} employed the PSD analysis to examine postural balance. This metric, as demonstrated in this study, can provide valuable insights into the dominant frequencies of sway. However, PSD analysis alone cannot provide a comprehensive understanding of the underlying neuromechanical strategies employed by individuals to maintain balance (see Fig.~\ref{PSD}), as it does not capture the spatial organization, coordination, or phase relationships between different body segments. 

Using the IP, multi-link inverted pendulum models, and the LQR controller, the current work analyzed the model's response to various conditions and provided the interpretation in terms of physiological and biomechanical implications. The LQR controller aligns with the hypothesis of optimal control in human neural balance systems~\cite{kuo1995optimal,Kooij1999,scott2004optimal} and provides a framework to interpret the physiological mechanisms underlying postural control. These mechanisms, governed by intricate neural feedback loops and coordinated muscle activation patterns across the ankle, knee, and hip joints, are reflected in the LQR parameters. For example, small $\beta_1$ values during stance indicated the ankle's dominant role in balance, suggesting heightened activation of ankle muscles and associated proprioceptive feedback systems. Conversely, large $\alpha$ values signified minimal neural and muscular activation, implying reliance on passive biomechanical properties for stability. The increase in $|\bar{\omega}^{\text{CF}}|$ under conditions such as elevated heights or visual perturbations revealed heightened neural and muscular engagement. This physiological response manifested as increased muscle co-contraction and enhanced sensory integration to maintain balance. Notably, the exoskeleton's influence on these physiological responses was evident in the reduced $|\bar{\omega}^{\text{CF}}|$ and higher $\alpha$ values, indicating a shift towards passive control strategies. This shift suggests decreased muscle activation, particularly at the knee joint. Physiologically, this translates to reduced muscular fatigue and improved postural stability over prolonged use.

The real-world application of the findings in this study is twofold. First, the results in this work demonstrate that exoskeletons enhance balance and reduce postural sway, particularly in environments such as high elevations or unstable floor surfaces. This suggests that exoskeletons play a significant role in reducing fall risks and improving worker safety. Moreover, enhanced postural stability in skilled tasks such as welding implies that exoskeletons can also contribute to reduced work errors and lead to improved productivity. Additionally, the research outcomes provide exoskeleton designers insights into joint-level balance strategies. This helps develop personalized control algorithms that respond to a worker’s changing balance strategy and tailor the device design to individual users. The method can also be integrated into VR/MR-based training programs, allowing workers to safely practice and improve their balance strategies in a controlled environment before facing real-world challenges. Furthermore, construction companies could adopt the methods for risk assessment, identifying potential fall risks associated with specific tasks or environments. Finally, the analysis presented in this work can serve as a tool for performance monitoring and allow timely interventions to enhance worker safety.

Despite the reported benefits in terms of stability and balance, some subjects expressed that wearing the knee exoskeleton was less comfortable during kneeling compared to the stance posture. This disparity in comfort levels may be attributed to the unique biomechanical demands and joint configurations associated with kneeling. These observations underscore the importance of considering posture-specific design factors and adaptability when developing wearable exoskeletons for construction workers. There are some other limitations to consider for future research directions. First, this study only recruited young, healthy students rather than experienced construction workers, it's important to acknowledge that experienced workers may have developed specific strategies or physical adaptations that could influence their interaction with exoskeletons. Expanding the subject pool to include experienced construction workers of varying ages will also be important to comprehensively assess the exoskeleton's effects on balance and performance. Second, incorporating direct physiological measurements, such as electromyography (EMG), could provide deeper insights into muscle activation and neuromuscular responses. This would allow for a detailed analysis of how exoskeletons reduce physical workload on key joints. Finally, investigating fatigue effects and developing adaptive control systems that respond to real-time balance shifts would further enhance the long-term safety, safety, and efficacy of exoskeletons in construction environments.

\section{Conclusions}
\label{con}

This study provided insights into the influence of lower-limb joints on neural balance control in quiet stance and kneeling gaits in construction. The use of triple- and double-link inverted pendulum models, coupled with the intersection point height frequency analysis and the linear quadratic controller, allowed for a comprehensive evaluation of the balance strategies of the subject at the joint level. The integration of VR/MR tools to simulate elevated environments and welding tasks enabled the assessment of the impact of elevation and wearable knee exoskeletons on postural balance and skilled task performance. The multi-subject experiment results highlight the critical role of knee joint strategies in maintaining postural stability and the potential of wearable exoskeletons in enhancing worker safety and performance in hazardous construction environments. The evaluation pipeline developed in this study can be further applied to assessing the potential of wearable assistive devices across diverse occupational work settings. 

\bibliographystyle{IEEEtran}
\bibliography{Ref}

\end{document}